\documentclass{article}

\usepackage[preprint]{corl_2026} 

\usepackage[utf8]{inputenc} 
\usepackage[T1]{fontenc}    
\usepackage{hyperref}       
\usepackage{url}            
\usepackage{booktabs}       
\usepackage{amsfonts}       
\usepackage{nicefrac}       
\usepackage{microtype}      
\usepackage{xcolor}         
\usepackage{graphicx}

\usepackage{algorithm}
\usepackage[noend]{algpseudocode}
\algrenewcommand\algorithmiccomment[1]{\hfill\small\textcolor{gray}{\textit{// #1}}}
\usepackage{booktabs}
\usepackage{multirow}
\usepackage{colortbl,xcolor}
\usepackage{soul}
\colorlet{llgray}{lightgray!40}
\sethlcolor{llgray}
\usepackage{mathtools}
\usepackage{caption}
\usepackage{wrapfig}
\usepackage{subfigure}
\usepackage{subcaption}

\usepackage{placeins}  
\usepackage{float}     
\usepackage{etoolbox}  

\usepackage{tabularx}
\usepackage{array}
\usepackage{makecell}

\newcolumntype{Y}{>{\raggedright\arraybackslash}X}
\newcolumntype{C}{>{\centering\arraybackslash}X}

\usepackage{amsmath,amsfonts,bm,amsthm}

\def\1{\bm{1}}

\DeclareMathAlphabet{\mathsfit}{\encodingdefault}{\sfdefault}{m}{sl}
\SetMathAlphabet{\mathsfit}{bold}{\encodingdefault}{\sfdefault}{bx}{n}

\title{RoboFFT: Finetuning generative robot policy via online reinforcement learning with forward process}

\author{
    Yu Li$^{*123}$, 
    Shenghe Hu$^{*34}$, 
    Yuhan Wang$^{3}$,  \\
    \textbf{Yaoxiang Pu}$^{3}$, 
    \textbf{Haotong Zhang}$^{3}$, 
    \textbf{Yuanpei Chen}$^{\dagger3}$, 
    \textbf{Yaodong Yang}$^{\dagger13}$ \\
}

\begin{document}
\maketitle

\begin{figure}[!h]
  \centering
  \vspace{-1.0em}

      \includegraphics[width=1.0\textwidth]{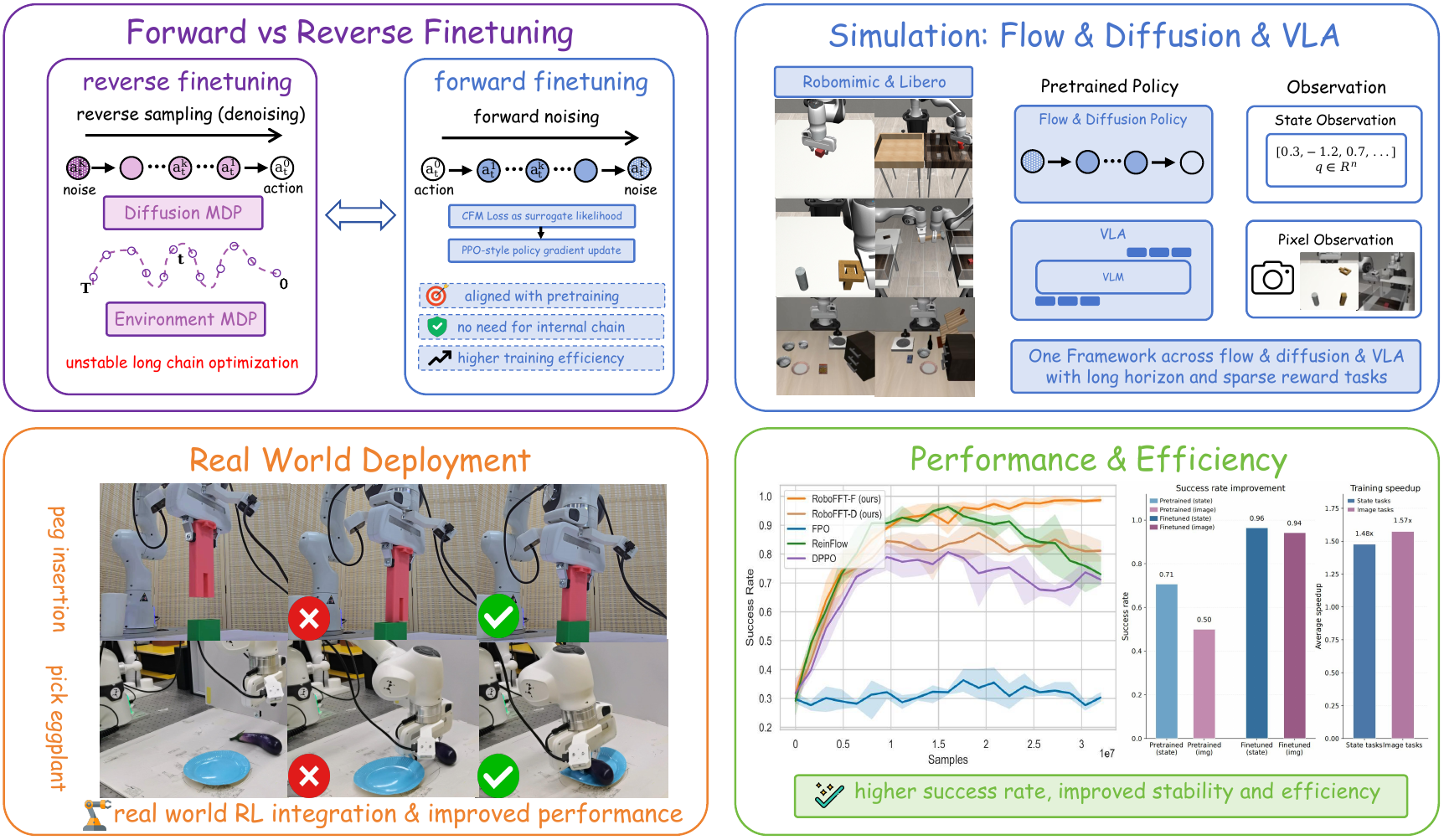}
  \caption{
  We present \textbf{RoboFFT}, an online reinforcement learning framework for finetuning generative robot policy with forward process. (a) Reverse finetuning formulates the reverse denoising process as a multi-step MDP, while forward finetuning aligns with the same objective as training by following the forward noising process. 
  (b) One framework across popular generative robot policy on simulation benchmarks. 
  (c) Real-world deployment with integration of  real world RL.
  (d) Improved performance in success rate, training stability and efficiency. 
  }
  \label{fig:teasor}
\end{figure}


\renewcommand{\thefootnote}{}
\footnotetext{%
\raggedright
\textsuperscript{1}\!Institute for Artificial Intelligence, Peking University.\;
\textsuperscript{2}\!Zhongguancun Academy.\;
\textsuperscript{3}\!PKU-PsiBot Joint Lab.\;
\textsuperscript{4}\!Nanjing University.\;
\textsuperscript{*}\!Equal contribution.\;
Author Email: \texttt{<yuanpei.chen312@gmail.com, yaodong.yang@pku.edu.cn>}. 
\textsuperscript{\textdagger}\!Corresponding author.
}


\begin{abstract}
Generative models, such as diffusion and flow-based models, have shown strong promise for robot policy learning by capturing complex and multimodal action distributions from demonstrations. However, policies trained solely with imitation learning often suffer from imperfect demonstrations and distributional shifts, while further improvement typically requires additional expert data. 
Reinforcement learning offers a natural solution through environment interaction, but effectively finetuning generative robot policies remains challenging due to the intractability of likelihood estimation. 
In this work, we propose \textbf{RoboFFT}, a forward-process reinforcement learning framework for finetuning generative robot policies, which applies forward noising to sampled actions and uses the weighted score / flow matching loss to construct a surrogate policy ratio for PPO-style updates. 
We evaluate \textbf{RoboFFT} with popular generative robot policies on representative simulation benchmarks, including long-horizon planning and sparse reward settings. Extensive experiments and analysis demonstrate that \textbf{RoboFFT} consistently improves performance while achieving better stability and training efficiency. We further integrate \textbf{RoboFFT} into a real world RL framework and demonstrate its effectiveness in real world tasks. Project website: \href{https://student-of-holmes.github.io/RoboFFT/}{https://student-of-holmes.github.io/RoboFFT/}.
\end{abstract}

\keywords{Diffusion RL, Generative Robot Policy, Robot Manipulation}


\section{Introduction}
\label{sec:intro}
Generative robot policies, typically based on diffusion and flow-based models \citep{song2020score, song2020denoising, ho2020denoising, lipman2022flow}, have become a powerful paradigm for learning visuomotor skills from demonstrations by modeling complex and multi-modal action distributions~\cite{chi2025diffusion, ze20243d, yan2025maniflow}, which significantly improves robot action accuracy and robustness and yields substantial performance gains in complex manipulation tasks~\citep{chi2025diffusion, yan2025maniflow}. However, policies trained purely by imitation learning remain constrained by the quality and coverage of the demonstration data. Imperfect demonstrations, covariate shift, and insufficient exposure to deployment-time states can all limit the final task performance of behavior-cloned policies~\cite{codevilla2019exploring, sagheb2025counterfactual}. Online reinforcement learning (RL) naturally presents itself as a promising paradigm to further improve generative robot policies, as it empowers the policy to directly maximize task rewards by continuously interacting with and learning from the environment. 


Applying policy gradient RL methods to generative robot policies is nevertheless nontrivial. Unlike autoregressive policies, whose token likelihoods are tractable, diffusion and flow-based models typically lack a tractable exact likelihood formulation. This makes popular policy gradient objectives, such as PPO, difficult to be applied directly. Existing approaches address this issue by formulating the reverse denoising process as a multi-step MDP embedded in the high-level environment MDP, and optimizing through this two-level decision making trajectory~\cite{black2023training, ren2024diffusion, zhang2025reinflow} (Fig.~\ref{fig:teasor}(a), left). Despite its promising results, reverse-process finetuning introduces substantial computation and memory overhead, and may suffer from gradient instability due to long denoising chains, especially in high-dimensional and long-horizon manipulation tasks.

To address these issues, we propose \textbf{RoboFFT}, a simpler yet more effective framework for online RL finetuning of generative robot manipulation policies through the \emph{forward} noising process, inspired by recent studies from other application domains such as image generation and simple locomotion \citep{xue2025advantage, zheng2025diffusionnft, mcallister2025flow}. Instead of assigning credit through the reverse denoising chain, \textbf{RoboFFT} approximates the exact log-likelihood with a more computationally stable and efficient form akin to the score / flow matching loss for pretraining, yielding a principled PPO-style objective that is aligned with the imitation-learning objective while avoiding complicated likelihood computation or backpropagation through the full reverse trajectory. For flow-based policies, we further adopt SDE sampling by injecting noise into the sampling process, enabling controllable exploration during online data collection. In practice, Monte Carlo estimation, ratio-temperature calibration and further adaptations make likelihood estimation and updates robust across different policy architectures and tasks.


We evaluate \textbf{RoboFFT} on challenging robot manipulation benchmarks such as Robomimic~\citep{mandlekar2021matters} and LIBERO~\citep{liu2023libero}, which pose difficulties like high-dimensional visual observation, long horizon planning and sparse reward for finetuning. Experimental results across different tasks demonstrate substantial and consistent performance improvements over pretrained diffusion and flow policies and achieve competitive or better performance compared with representative reverse and forward process finetuning baselines, while reducing optimization overhead. We further verify the applicability of our method on more complicated VLA and real world deployment settings, and study the key designs through ablations and visualizations. Our contributions are summarized as follows:

(1) We introduce \textbf{RoboFFT}, an online reinforcement learning framework that performs policy gradient finetuning in the forward noising process for generative robot policies.

(2) We carry out practical implementation of \textbf{RoboFFT} for representative generative robot policies (flow, diffusion and VLA) on challenging long-horizon, sparse-reward manipulation tasks, showing that \textbf{RoboFFT} achieves superior performance and improved training stability and efficiency. 

(3) We integrate \textbf{RoboFFT} into the real world reinforcement learning framework for practical deployment and demonstrate its effectiveness over real world tasks. 

(4) We carry out a systematic ablation analysis of key design choices and conduct investigative experiments to provide empirical insights for the mechanism of \textbf{RoboFFT}.

\section{Related Work}
\label{sec:related}

\subsection{Policy Gradient for Robot Manipulation}

Policy gradient and actor critic methods provide a standard framework for learning continuous-control policies from interaction data~\citep{williams1992simple,sutton1999policy,lillicrap2015continuous,schulman2017proximal}. Due to the
training stability with high-dimensional action spaces, they have been widely applied to a range of robotic control problems, including dexterous manipulation, legged locomotion, and agile navigation~\citep{wang2025dexterous,chen2022system,chen2022towards,andrychowicz2020learning,hwangbo2019learning,kaufmann2023champion}. For long-horizon manipulation, however, training from scratch remains highly exploration-intensive and is often impractical. A more practical paradigm is therefore to finetune policies initialized from demonstration data. Prior studies have shown that reinforcement learning can further improve such demonstration-initialized policies beyond behavior cloning while substantially reducing the exploration burden~\citep{rajeswaran2017learning,torne2024reconciling,peng2021amp}. RoboFFT follows this finetuning paradigm, while focusing on diffusion- and flow-based robot policies whose action distributions are implicitly induced by generative sampling processes, making the likelihood ratio computation required by policy gradient methods nontrivial.


\subsection{RL for Generative Models}

Reinforcement learning has recently been widely applied to generative model finetuning in offline~\citep{ball2023efficient,nakamoto2024cal,hu2023imitation, wang2022diffusion, hansen2023idql, park2025flow, celik2025dime}, offline-to-online~\citep{chen2025conrft,lei2025rl}, and online settings~\citep{ren2024diffusion,zhang2025reinflow,yi2026flow,mcallister2025flow,xue2025advantage,zheng2025diffusionnft}. Online finetuning for diffusion and flow models can be broadly categorized into reverse-process and forward-process approaches. Reverse-process methods formulate the denoising or sampling trajectory as a multi-step decision process, enabling tractable per-step likelihood for policy optimization~\citep{black2023training,ren2024diffusion,zhang2025reinflow}. While effective, optimizing through long internal sampling chains can increase computation and memory costs and may introduce training instability. Forward-process methods instead optimize policy via constructing surrogate likelihood objectives or contrastive learning, better aligning RL finetuning with generative pretraining~\citep{xue2025advantage,mcallister2025flow,yi2026flow,zheng2025diffusionnft}. DiffusionNFT and AWM~\citep{zheng2025diffusionnft, xue2025advantage} study this idea for text-to-image generation. FPO and FPO++~\citep{mcallister2025flow,yi2026flow} extend it to robotic control, but mainly focus on locomotion tasks. In comparison, RoboFFT studies forward-process finetuning in more challenging robotic manipulation, and covers more extensive settings including both diffusion and flow policies, VLA finetuning, and real world deployment with practical stabilization designs.

\section{Preliminaries}
\label{sec:preliminaries}
\subsection{Policy Gradient and PPO}
\label{subsec:prelim_pg_and_ppo}

We consider an observation conditioned robot policy $\pi_{\theta}(a_t|o_t)$,
where $a_t$ can denote either a single action or an action chunk. Given buffer $\mathcal{B}$ of rollout data collected by the previous policy
$\pi_{\theta_{\mathrm{old}}}$, PPO optimizes the clipped likelihood-ratio
objective which takes expectation over $(o_t,a_t)\sim \mathcal{B}$:
\begin{equation}
\begin{aligned}
    J_{\mathrm{PPO}}(\theta)
    =
    \mathbb{E}
    \left[
    \min \left(
        \rho_t(\theta)\hat{A}_t,\,
        \mathrm{clip}
        \left(
            \rho_t(\theta),
            1-\epsilon_{\mathrm{clip}},
            1+\epsilon_{\mathrm{clip}}
        \right)
        \hat{A}_t
    \right)
    \right], \ 
    \rho_t(\theta)=\frac{\pi_{\theta}(a_t|o_t)}{\pi_{\theta_{\mathrm{old}}}(a_t|o_t)}
\end{aligned}
\label{eq:ppo}
\end{equation}
where $\rho_t(\theta)$ is the likelihood ratio and $\hat{A}_t$ denotes the estimated advantage. This update requires evaluating the policy ratio on rollout
actions, which is straightforward for explicit policies such as Gaussians,
but generally nontrivial for diffusion- and flow-based robot policies whose
action distributions are implicitly induced by iterative ODE / SDE samplers.

\subsection{Diffusion- and Flow-Based Robot Policies}
\label{subsec:prelim_diffusion_and_flow}
A diffusion- or flow-based robot policy generates a clean action or action chunk
$a_t^0$ conditioned on the observation $o_t$. During imitation pretraining, the
demonstration action is perturbed through a forward noising path
\begin{equation}
    a_t^k
    =
    \alpha_k a_t^0 + \sigma_k \epsilon,
    \qquad
    \epsilon \sim \mathcal{N}(0,I),
    \qquad
    k\in[0,1],
\end{equation}
where $\alpha_k$ and $\sigma_k$ define the noise schedule. The policy models the action distribution by learning a score function or velocity field
$u_{\theta}(a_t^k,k,o_t)$ with the score / flow matching loss, which can be generally written as

\begin{equation}
\begin{aligned}
    \mathcal{L}_{\theta}^{w}(a_t^0,o_t)
    =
    \mathbb{E}_{k\sim\mathcal U(0,1),\epsilon \sim\mathcal{N}(0,I)}
    \left[
        w(\lambda_k) \cdot\left(-\frac{\text{d}\lambda_k}{\text{d} k}\right)
        \left\|
            u_{\theta}(a_t^k,k,o_t)
            -
            y_k(a_t^0,\epsilon)
        \right\|_2^2
    \right].
\end{aligned}
\end{equation}
where $\lambda_k=\log(\alpha_k^2/\sigma_k^2)$ denotes the log-SNR schedule, and $w(\lambda_k)$ is a SNR-dependent weight ~\citep{kingma2023understanding}. The target $y_k$ specifies different policy parameterizations: for noise-predicting diffusion policies, the noise network $u_\theta=\epsilon_\theta$ is regressed towards the target noise $y_k(a_t^0,\epsilon)=\epsilon$. For flow-based policies, $u_\theta=v_\theta$ and the target is the
path velocity $y_k(a_t^0,\epsilon)=v_k(a_t^0,\epsilon)=\dot{\alpha}_k a_t^0 + \dot{\sigma}_k \epsilon$, which reduces to $v_k(a_t^0,\epsilon)=\epsilon-a_t^0$ under the special case of rectified flow where $\alpha_k=1-k,\sigma_k=k$. Pretraining usually adopts $w$ s.t. $w(\lambda_k) \cdot\left(-\text{d}\lambda_k/\text{d} k\right)\equiv1$. At inference time, the learned $u_\theta$ induces the policy $\pi_{\theta}(a_t^0|o_t)$, a.k.a. marginal
distribution of the clean action sample through reverse denoising or flow
sampling, and thus the exact likelihood is not directly computable. RoboFFT
therefore constructs a surrogate policy ratio akin to the matching loss used for pretraining.

\begin{figure*}[!ht]
  \centering

      \includegraphics[width=1.0\textwidth]{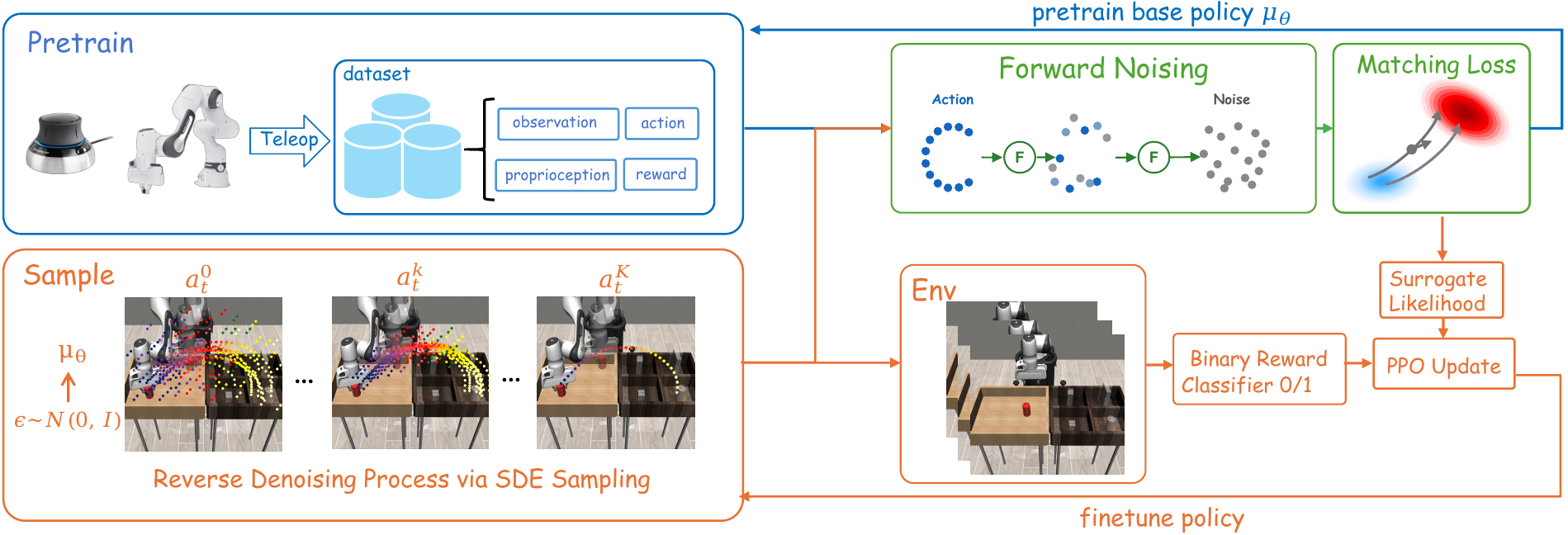}
  \caption{
  Overview of \textbf{RoboFFT}.
  We first collect the offline data leveraging teleoperation and pretrain the base policy with the score / flow matching loss. The pretrained policy then generates actions using an SDE sampler and receives reward feedback from the environment. Finally, we apply forward noising and reuse the matching loss as a surrogate likelihood for PPO-style policy update.
  }
  \label{fig:method}
\end{figure*}

\section{Method}
\label{sec:method}

In this section, we present \textbf{RoboFFT}, our online RL framework for fine-tuning pretrained generative robot policies, as shown in Fig.~\ref{fig:method}. Following Sec.~\ref{sec:preliminaries},
the policy is represented by an observation conditioned prediction network
$u_{\theta}(a_t^k,k,o_t)$ and pretrained with the score / flow matching loss
$\mathcal{L}_{\theta}^{w}(a_t^0,o_t)$. The main designs of our method are to introduce a stochastic sampling strategy that enhances data diversity and policy exploration (Sec.~\ref{subsec:odetosde}), and to formulate forward-process optimization, where the score / flow matching loss is employed as surrogate likelihood to enable principled policy gradient update (Sec.~\ref{subsec:surrogate_likelihood}). We additionally discuss some of the practical implementation details that help improving the training performance and stability (Sec.~\ref{subsec:practical_implementation}). The complete algorithmic description for RoboFFT is deferred to the appendix (Alg.~\ref{alg:RoboFFT}).

\subsection{SDE Sampling}
\label{subsec:odetosde}

We start from a policy pretrained on teleoperation data with the matching loss
defined in Sec.~\ref{sec:preliminaries}. During online rollouts, diffusion
policies naturally use their SDE reverse sampler. Flow-based policies, however, are typically sampled by a deterministic ODE sampler solving $\text{d}a^k_t = v_\theta(a^k_t,k) \, \text{d} k$, which lacks inherent stochasticity, thus not meeting the controlled stochastic exploration demand that PPO-style optimization typically benefits from. We therefore inject noise into the flow sampling process, turning it into an Euler-Maruyama SDE sampler with the following discretized form~\citep{kloeden1977numerical}:
\begin{equation}
\label{eq:sde_sampling}
    a_t^{k+\Delta k}
    =
    a_t^k
    +
    u_{\theta}(a_t^k,k,o_t)\Delta k
    +
    \sigma_{\mathrm{sam}}(k)\sqrt{\Delta k}\epsilon,
    \qquad
    \epsilon\sim\mathcal{N}(0,I).
\end{equation}
Though $\sigma_{\mathrm{sam}}(k)$ can depend on denoising timestep $k$, we simply set $\sigma_{\mathrm{sam}}(k)=\eta$ in practice, where $\eta>0$ is a fixed exploration strength. This sampler is used only for collecting rollout actions, unlike reverse-process methods where it also participates in the likelihood computation \cite{zhang2025reinflow}.

\subsection{Surrogate Likelihood for Policy Gradient}
\label{subsec:surrogate_likelihood}

For a rollout action $a_t^0$, the exact likelihood ratio
$\rho_t(\theta)$ is generally intractable as discussed in Sec.~\ref{subsec:prelim_diffusion_and_flow}. RoboFFT
therefore uses the matching loss $\mathcal{L}_{\theta}^{w}(a_t^0,o_t)$ as a surrogate negative
log-likelihood. This is motivated by the connection between weighted denoising
objectives and ELBOs~\citep{kingma2023understanding}, and follows recent
forward process policy gradient formulations~\citep{mcallister2025flow,yi2026flow}.
We defer the detailed ELBO interpretation to Appendix~\ref{app:method_details}.

Given the per action matching loss $\mathcal{L}_{\theta}^{w}(a_t^0,o_t)$, we
define the surrogate policy ratio as
\begin{equation}
\label{eq:surrogate_ratio}
    \widehat{\rho}_t(\theta)
    =
    \exp\left(
        \mathcal{L}_{\theta_{\mathrm{old}}}^{w}(a_t^0,o_t)
        -
        \mathcal{L}_{\theta}^{w}(a_t^0,o_t)
    \right).
\end{equation}
Replacing the exact PPO ratio in Eq.~\eqref{eq:ppo} with
$\widehat{\rho}_t(\theta)$ yields the RoboFFT policy objective:
\begin{equation}
\label{eq:robofft_objective}
\begin{aligned}
    J_{\mathrm{RoboFFT}}(\theta)
    =
    \mathbb{E}_{(o_t,a_t^0)\sim\mathcal{B}}
    \left[
    \min\left(
        \widehat{\rho}_t(\theta)\hat{A}_t,\,
        \mathrm{clip}
        \left(
            \widehat{\rho}_t(\theta),
            1-\epsilon_{\mathrm{clip}},
            1+\epsilon_{\mathrm{clip}}
        \right)
        \hat{A}_t
    \right)
    \right],
\end{aligned}
\end{equation}
where $\mathcal{B}$ is the rollout buffer and $\hat{A}_t$ is estimated with GAE.

\subsection{Practical Implementation}
\label{subsec:practical_implementation}
The following two main techniques are adopted to make the implementation of RoboFFT more stable.

\textbf{Monte Carlo sampling.}
The matching loss is an expectation over the forward noise level and Gaussian
perturbation. For each $a_t^0$, we estimate it with
$N_{\mathrm{mc}}$ i.i.d. samples to reduce the variance:
\begin{equation}
\label{eq:mc_matching_loss}
\begin{aligned}
    \widehat{\mathcal{L}}_{\theta}^{w}(a_t^0,o_t)
    =
    \frac{1}{N_{\mathrm{mc}}}
    \sum_{i=1}^{N_{\mathrm{mc}}}
    w(k_i)
    \left\|
        u_{\theta}(a_t^{k_i},k_i,o_t)
        -
        y_{k_i}(a_t^0,\epsilon_i)
    \right\|_2^2,
    \qquad
    a_t^{k_i}
    =
    \alpha_{k_i}a_t^0+\sigma_{k_i}\epsilon_i .
\end{aligned}
\end{equation}
Here $k_i\sim\mathcal{U}(0,1)$ and $\epsilon_i\sim\mathcal{N}(0,I)$. We reuse
the same $\{(k_i,\epsilon_i)\}$ for $\theta$ and $\theta_{\mathrm{old}}$ when
computing the loss difference. This helps stabilize policy gradient updates.

\textbf{Temperature scaling.}
Since the matching loss scale may deviate from the true log-likelihood scale due to surrogation, we 
introduce a ratio temperature $\alpha_{\mathrm{ratio}}$ to recalibrate it:
\begin{equation}
\label{eq:scaled_ratio}
    \widehat{\rho}^{\,s}_t(\theta)
    =
    \exp\left(
        \alpha_{\mathrm{ratio}}
        \left[
            \widehat{\mathcal{L}}_{\theta_{\mathrm{old}}}^{w}(a_t^0,o_t)
            -
            \widehat{\mathcal{L}}_{\theta}^{w}(a_t^0,o_t)
        \right]
    \right).
\end{equation}
Experimental results justify the necessity of all the designs above (Sec.~\ref{sec:abl_viz}). More experiment details and other techniques tailored for diffusion and VLA are deferred to Appendix~\ref{app:robomimic_impl} and~\ref{app:vla_details}.

\section{Experiments}
\label{sec:experiments}
In this section, we evaluate \textbf{RoboFFT} on simulated and real-world robot manipulation tasks. We first compare it with representative diffusion- and flow-policy RL baselines on Robomimic (Sec.~\ref{sec:baseline}), then test its applicability to a generalist VLA policy (Sec.~\ref{sec:vla}). We further analyze key design choices and policy behavior through ablation and visualization experiments (Sec.~\ref{sec:abl_viz}), and finally validate RoboFFT in real world manipulation tasks (Sec.~\ref{sec:real}).

\begin{figure*}[!ht]
  \centering

  \makebox[1.0\textwidth][c]{%
    \subfigure[Lift-state]{%
      \includegraphics[width=0.25\textwidth]{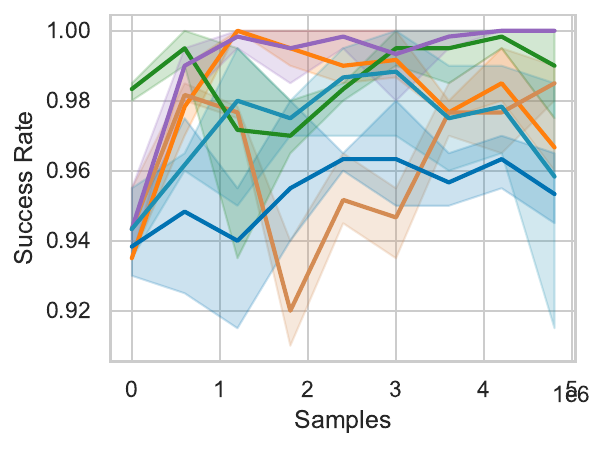}
    }
    \subfigure[Can-state]{%
      \includegraphics[width=0.25\textwidth]{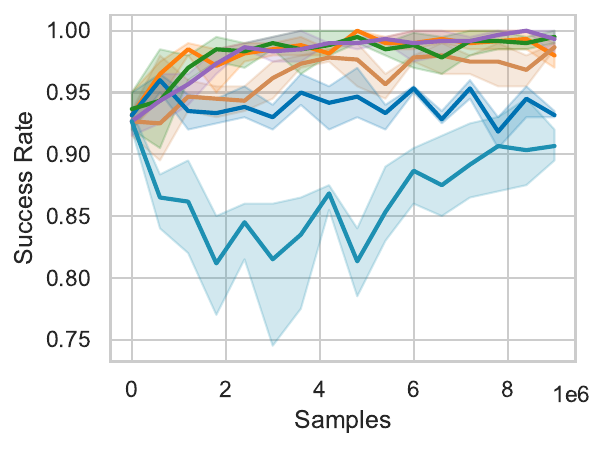}
    }
    \subfigure[Square-state]{%
      \includegraphics[width=0.25\textwidth]{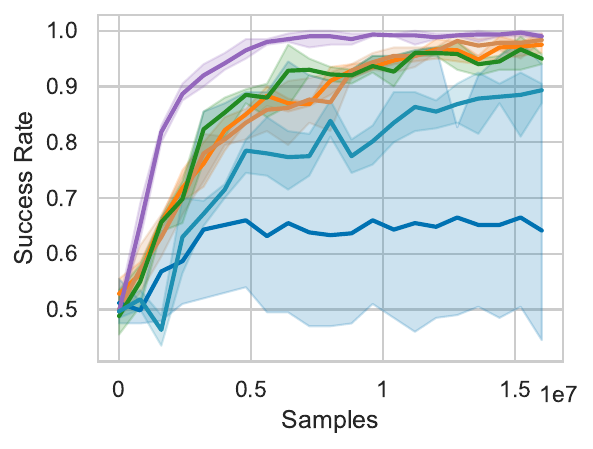}
    }
    \subfigure[Transport-state]{%
      \includegraphics[width=0.25\textwidth]{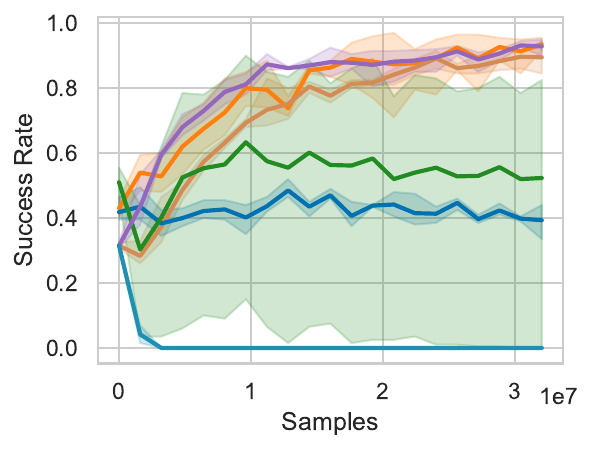}
    }
  }

  \makebox[1.0\textwidth][c]{%
    \subfigure[Lift-pixel]{%
      \includegraphics[width=0.25\textwidth]{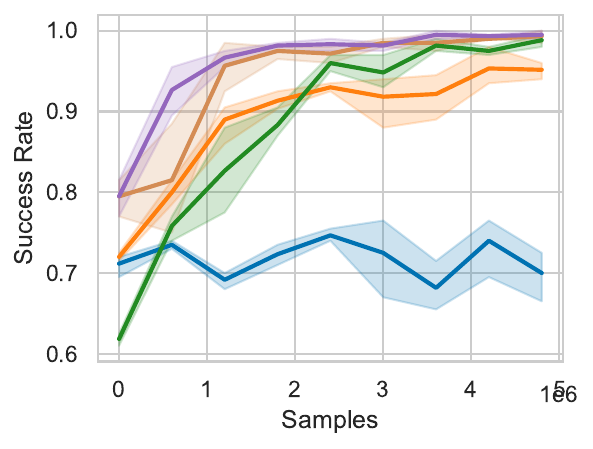}
    }
    \subfigure[Can-pixel]{%
      \includegraphics[width=0.25\textwidth]{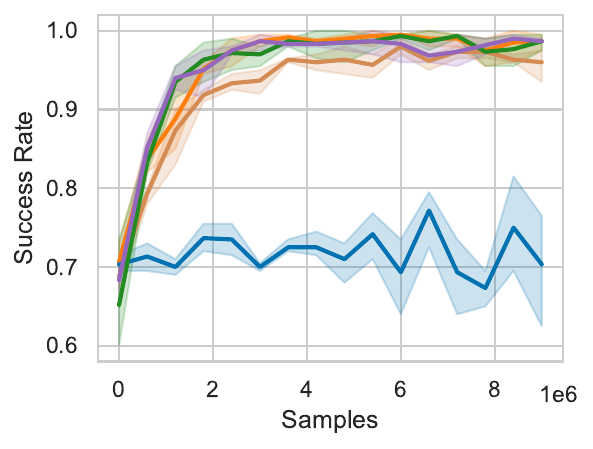}
    }
    \subfigure[Square-pixel]{%
      \includegraphics[width=0.25\textwidth]{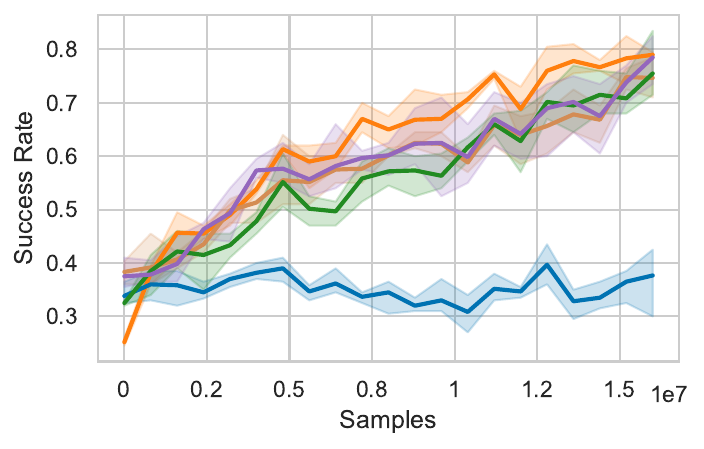}
    }
    \subfigure[Transport-pixel]{%
      \includegraphics[width=0.25\textwidth]{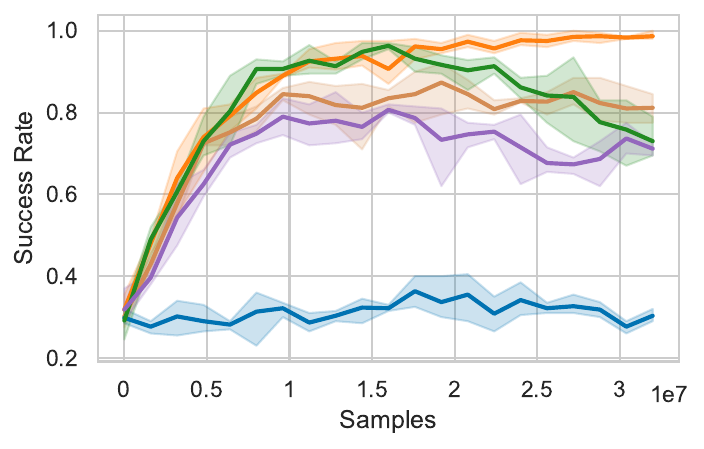}
    }
  }

  \vspace{0.4em}

  \includegraphics[width=0.75\textwidth]{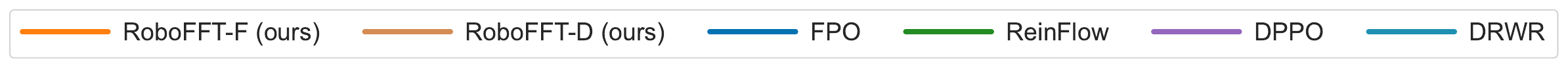}

  \caption{
  Success rates on Robomimic state- and pixel-input manipulation tasks. Curves show means over three random seeds, with shaded regions indicating one standard deviation across seeds. 
  }
  \label{fig:robomimic}
\end{figure*}

\begin{figure*}[!t]
  \centering
  \captionsetup{font=small,skip=2pt}

  \noindent
  \begin{minipage}[t]{0.485\textwidth}
    \centering
    \begin{minipage}[c][0.155\textheight][c]{\linewidth}
      \centering
      \includegraphics[
        width=0.98\linewidth,
        height=0.145\textheight,
        keepaspectratio
      ]{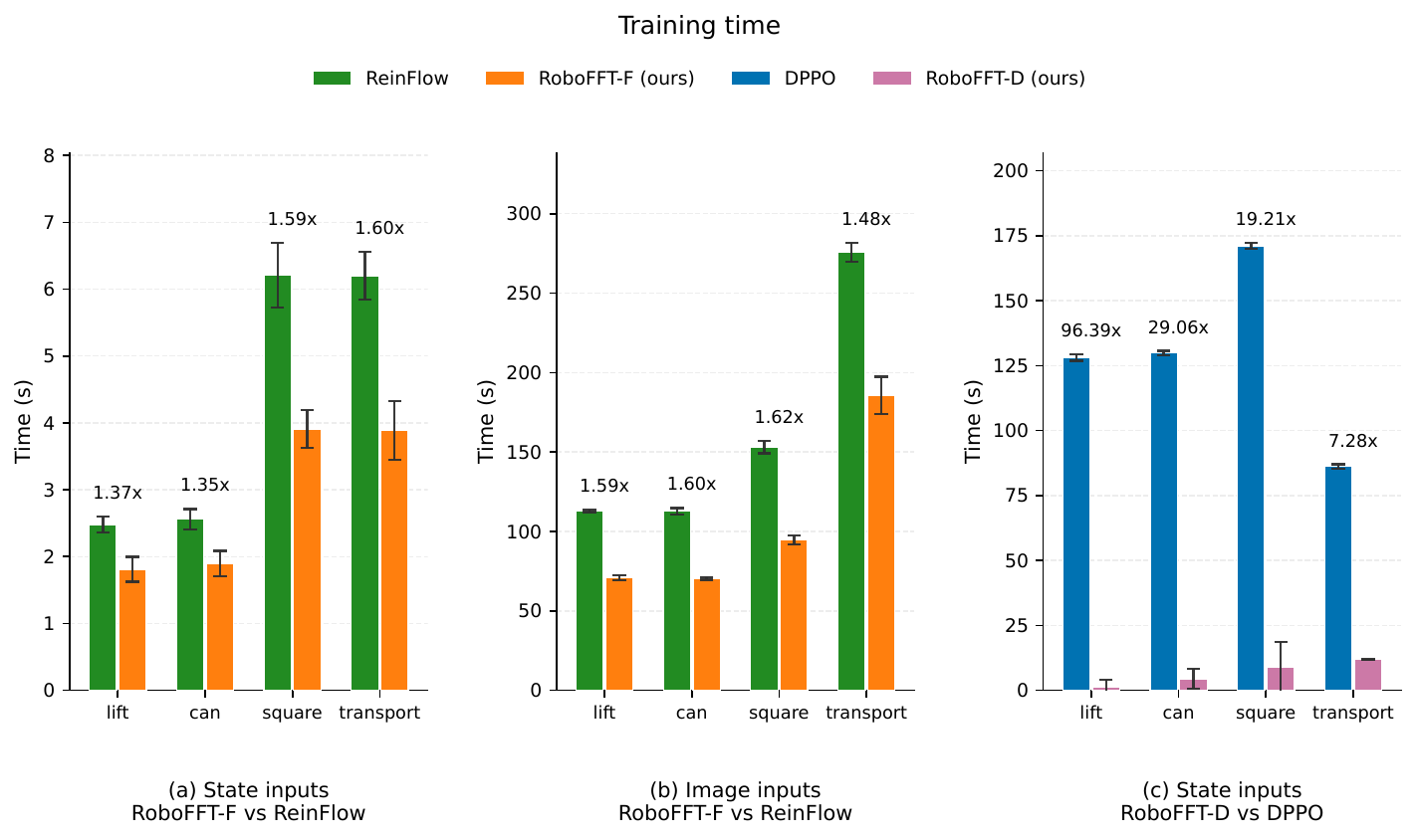} 
    \end{minipage}
  \end{minipage}
  \hfill
  \begin{minipage}[t]{0.485\textwidth}
    \centering
    \begin{minipage}[c][0.155\textheight][c]{\linewidth}
      \centering
      \includegraphics[
        width=0.98\linewidth,
        height=0.145\textheight,
        keepaspectratio
      ]{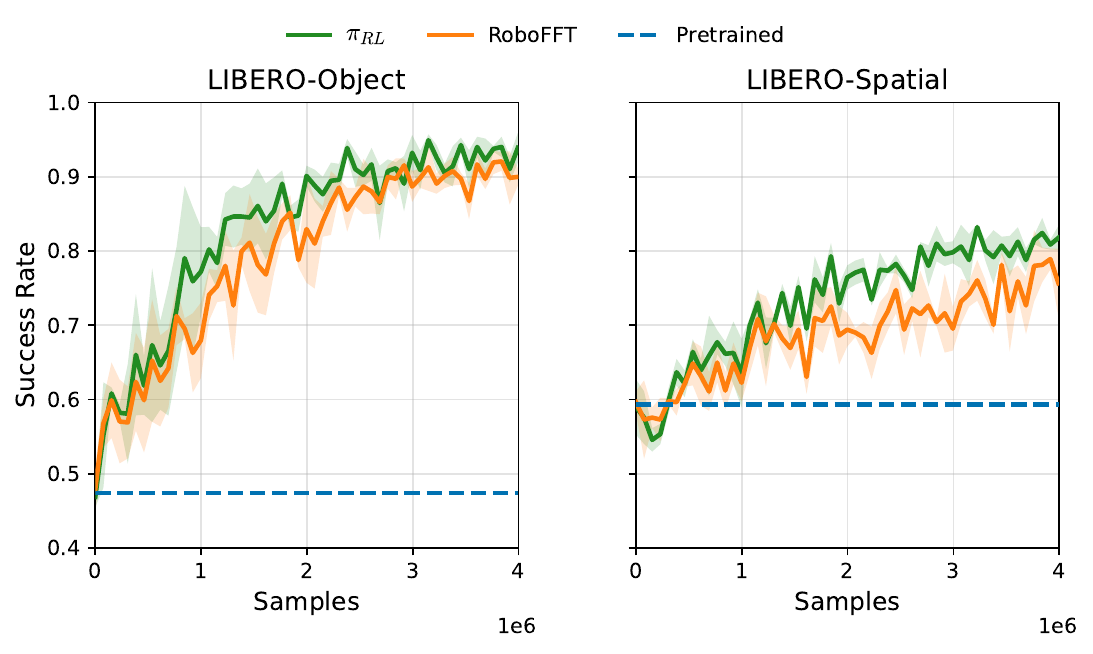}
    \end{minipage}
  \end{minipage}

  \vspace{-0.25em}

  \noindent
  \begin{minipage}[t]{0.485\textwidth}
    \captionof{figure}{
        Wall-clock training time for RoboFFT-F versus ReinFlow and
        RoboFFT-D versus DPPO. Bars show means and error bars show standard deviations over the first 10 training iterations of each configuration. Training time includes policy and value-network updates and excludes rollout collection.
    }
    \label{fig:time_costs_training} 
  \end{minipage}
  \hfill
  \begin{minipage}[t]{0.485\textwidth}
    \captionof{figure}{
    VLA finetuning performance on LIBERO-Object and LIBERO-Spatial. RoboFFT consistently improves the pretrained policy across both task suites and achieves competitive performance compared with $\pi_{\mathrm{RL}}$, demonstrating its effectiveness for finetuning generalist robot policy like VLA.
    }
    \label{fig:vla_finetuning}
  \end{minipage}

  \vspace{-6mm}
\end{figure*}


\subsection{Comparison to flow and diffusion-based RL algorithms}
\label{sec:baseline}




We instantiate RoboFFT with generative robot policies including diffusion (\textbf{RoboFFT-D}) and rectified flow (\textbf{RoboFFT-F}), and compare them against representative online finetuning baselines. DPPO~\citep{ren2024diffusion} and ReinFlow~\citep{zhang2025reinflow} optimize through the reverse sampling process, while DRWR and FPO~\citep{mcallister2025flow,yi2026flow} follow forward-process finetuning. DRWR uses reward-weighted regression, and FPO applies a forward-process PPO-style ratio but is not specifically adapted to the high-dimensional, long-horizon robot manipulation setting. We evaluate all methods on four sparse-reward Robomimic tasks, \textit{Lift}, \textit{Can}, \textit{Square}, and \textit{Transport}, under both state and pixel observations.
As shown in Fig.~\ref{fig:robomimic}, RoboFFT achieves finetuning performance competitive with or better than reverse-process baselines. On easier tasks such as \textit{Lift} and \textit{Can}, RoboFFT reaches comparable final success rates. On the more challenging \textit{Square} and \textit{Transport} tasks, especially under pixel observations, RoboFFT, particularly RoboFFT-F which features rectified sampling, shows more stable improvement and stronger performance. In contrast, DRWR and FPO are often less competitive, likely due to the lack of manipulation-oriented enhancements. These suggest that RoboFFT retains the simplicity of forward-process optimization while scaling better to visual, long-horizon, and dexterous manipulation tasks. To better reveal the stability of our method, additional evaluations with more random
seeds are provided in Appendix~\ref{app:additional_seeds}.

To examine the efficiency of our method, we further compare per-iteration wall-clock training times within each policy family: RoboFFT-F against ReinFlow, and RoboFFT-D against DPPO (Fig.~\ref{fig:time_costs_training}). RoboFFT-F achieves an average training-time speedup of $1.53\times$ over ReinFlow, while on the four Robomimic state-input tasks RoboFFT-D achieves an average training-time speedup of $38.0\times$ against DPPO, despite using $N_{\mathrm{mc}}=16$. Thus, the computational advantage of the forward-process update over the reverse-process likelihood chains persists for both flow and diffusion policies, even with multiple Monte Carlo evaluations. The measurement protocol and aggregation of task-wise speedups are detailed in Appendix~\ref{app:efficiency_analysis}.


\subsection{Evaluation on generalist robot policy}
\label{sec:vla}

To further investigate the scalability of \textbf{RoboFFT} to more sophisticated policy parameterizations, we conduct experiments on the generalist robot policy $\pi_0$, a state-of-the-art vision-language-action (VLA) model~\citep{black2024pi, intelligence2025pi}. $\pi_0$ contains 3.3B parameters and adopts a VLM-based backbone together with a flow-based action head for continuous action generation. This architecture makes it compatible with \textbf{RoboFFT}, as the proposed forward process policy optimization can be directly applied to the flow matching objective of the action head.
Nevertheless, finetuning $\pi_0$ is substantially more challenging than finetuning task-specific generative policies. In particular, $\pi_0$ predicts long horizon action chunks with high-dimensional action representations. Directly summing the flow matching loss over such a large action space can lead to large loss magnitudes and unstable policy updates, which poses a great challenge for scaling \textbf{RoboFFT} to VLA.
We evaluate \textbf{RoboFFT} on the \textit{Spatial} and \textit{Object} tasks of the \textit{LIBERO} benchmark. As shown in Fig.~\ref{fig:vla_finetuning}, \textbf{RoboFFT} consistently improves the performance of the pretrained $\pi_0$ policy and achieves competitive results compared with the $\pi_{\mathrm{RL}}$ baseline~\citep{chen2025pirl}. These results demonstrate that the proposed forward-process finetuning framework can scale to large VLA policies while maintaining effective and stable policy improvement.

\begin{figure*}[!th]
  \vspace{-2mm}
  \centering
  \captionsetup{font=small,skip=2pt}

  \noindent
  \begin{minipage}[t]{0.485\textwidth}
    \centering
    \begin{minipage}[c][0.145\textheight][c]{\linewidth}
      \centering
      \begin{minipage}[c]{0.48\linewidth}
        \centering
        \includegraphics[
          width=\linewidth,
          height=0.115\textheight,
          keepaspectratio
        ]{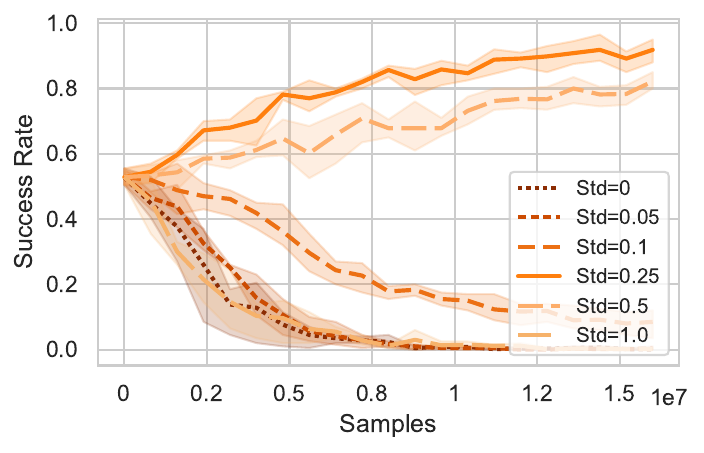}\\[-0.15em]
        {\small (a) Noise level}
        \label{fig:ablation_std}
      \end{minipage}
      \hfill
      \begin{minipage}[c]{0.48\linewidth}
        \centering
        \includegraphics[
          width=\linewidth,
          height=0.115\textheight,
          keepaspectratio
        ]{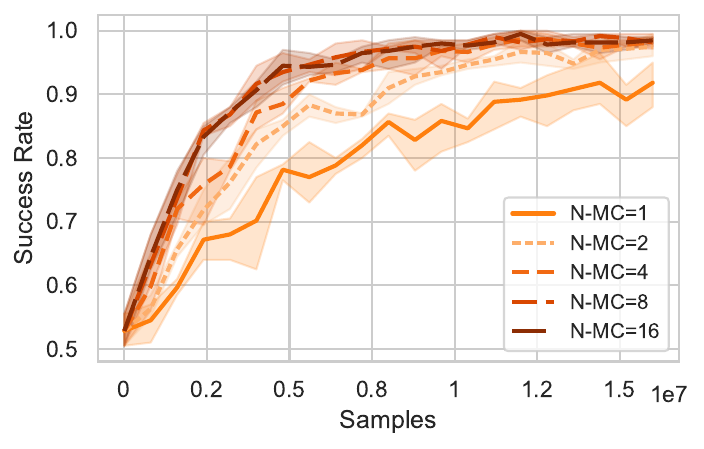}\\[-0.15em]
        {\small (b) MC repetition}
        \label{fig:ablation_nmc}
      \end{minipage}
    \end{minipage}
  \end{minipage}
  \hfill
  \begin{minipage}[t]{0.485\textwidth}
    \centering
    \begin{minipage}[c][0.145\textheight][c]{\linewidth}
      \centering
      \includegraphics[
        width=0.98\linewidth,
        height=0.128\textheight,
        keepaspectratio
      ]{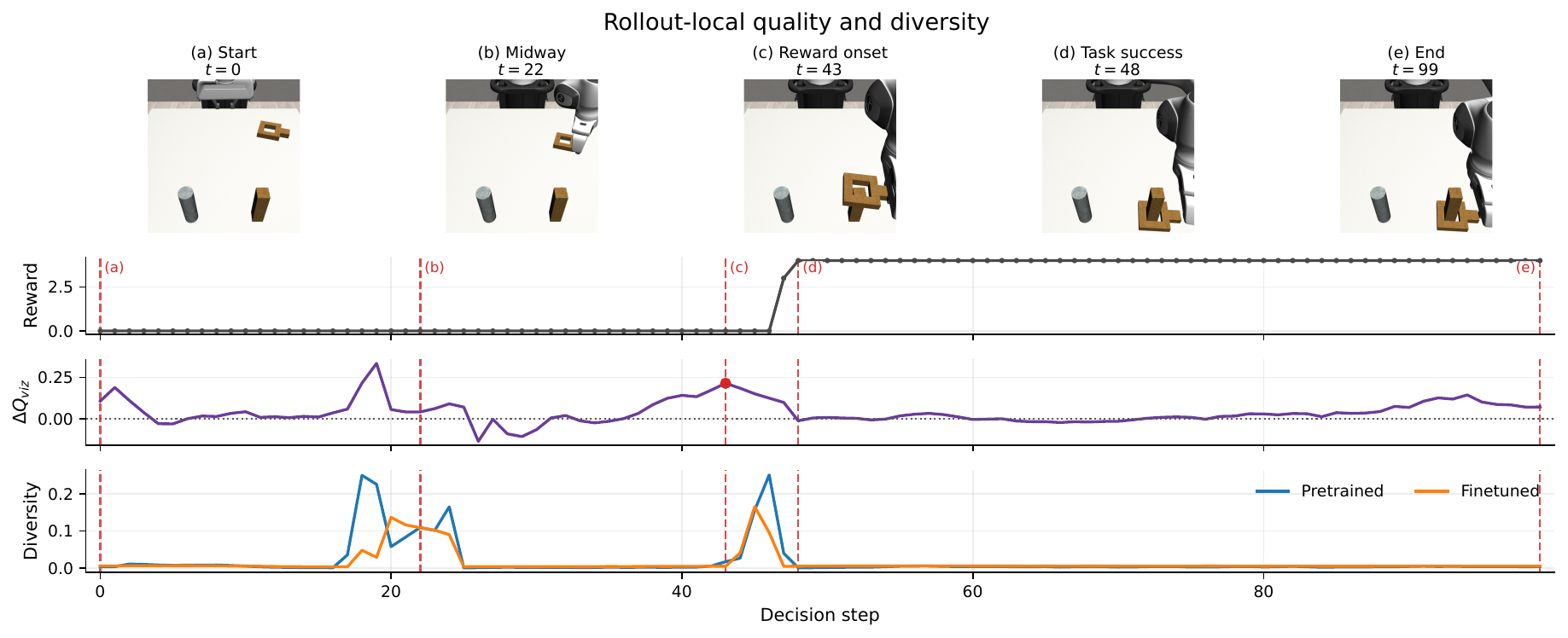}
    \end{minipage}
  \end{minipage}

  \vspace{0.15em}

  \noindent
  \begin{minipage}[t]{0.485\textwidth}
    \captionof{figure}{
    Key ablations on Robomimic Square-state. Moderate SDE sampling noise improves exploration, while a small number of Monte Carlo samples is sufficient for stable surrogate-ratio estimation.
    }
    \label{fig:ablation_row}
  \end{minipage}
  \hfill
  \begin{minipage}[t]{0.485\textwidth}
    \captionof{figure}{
    Rollout-level visualization on Robomimic Square. The finetuned policy increases diagnostic action quality while reducing action diversity around key decision stages.
    }
    \label{fig:viz_rollout}
  \end{minipage}

  \vspace{-2mm}
\end{figure*}

\subsection{Ablation and visualization experiments}
\label{sec:abl_viz}

\noindent\textbf{Ablation studies.} We perform ablation analysis on two main design choices with RoboFFT-F on Robomimic Square-state, a sparse-reward task with representative manipulation difficulty. Fig.~\ref{fig:ablation_row}(a) varies the SDE sampling noise level $\eta$ in Eq.~\eqref{eq:sde_sampling}, which shows that moderate SDE sampling noise provides the best balance between exploration and stable policy improvement, while insufficient exploration leads to policy collapse and overly large noise also degrades performance. Fig.~\ref{fig:ablation_row}(b) studies the MC sampling repetition $N_{\mathrm{mc}}$ used to estimate the surrogate matching loss in Eq.~\eqref{eq:mc_matching_loss}. Increasing $N_{\mathrm{mc}}$ from $1$ to $2$ accelerates convergence, while further increases yield limited gains within the online interaction budget. This supports using a small MC budget in this flow-policy setting, rather than implying uniformly low estimation variance across generative policies. Broader analyses in Appendix~\ref{app:ablation_details} examine secondary factors for RoboFFT-F, perform additional ablations for RoboFFT-F on Transport-state, and ablate unique factors for RoboFFT-D, examining their roles across tasks and policy parameterizations.

\begin{figure}[!h]
  \centering

  \begin{minipage}[b]{0.45\textwidth}
  \includegraphics[width=\textwidth]
    {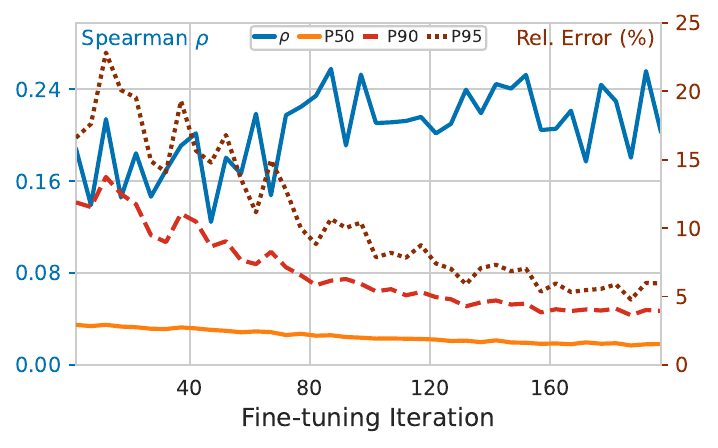}
  \vspace{-6mm}
  \caption{
  Surrogate-ratio discrepancy during RoboFFT-F finetuning on
  Robomimic Square-state, represented by Spearman rank correlation and relative error quantiles.
  }
  \label{fig:surrogate_ratio_gap}
  \end{minipage}
  \hfill
  \begin{minipage}[b]{0.48\textwidth}
    \centering
    \small
    \setlength{\tabcolsep}{1.5pt}
    \begin{tabular}{lcccc}
    \toprule
    Task
    & Pretrain
    & Offline
    & Online
    \\
    \midrule
    pick eggplant 
    & 12 / 20
    & 18 / 20
    & 20 / 20
    \\
    peg insertion
    & 10 / 20
    & 19 / 20
    & 20 / 20
    \\
    Pour
    & 9 / 20
    & 15 / 20
    & 18 / 20
     \\
    cabinet retrieval
    & 5 / 20
    & 14 / 20
    & 17 / 20
     \\
    \toprule
    Task
    & BC Agg.
    & DRWR
    & RL100
    \\
    \midrule
    pick eggplant 
    & 14 / 20
    & 12 / 20
    & 20 / 20
    \\
    \bottomrule
    
    \end{tabular}
    \caption{Real-world success rates. Top: our method exhibits consistent improvement of four tasks from Pretrain to Offline to Online. Bottom: our method outperforms baselines.}
    \label{tab:real_world}
  \end{minipage}
\end{figure}

\noindent\textbf{Surrogate-ratio discrepancy.} To understand how the matching loss serves as an effective surrogate of the log likelihood, we quantify its discrepancy from
a ReinFlow-style closed-form reference during RoboFFT-F finetuning
on Square-state. Using 2,048 action chunks per iteration, we measure the Spearman rank correlation and quantiles of relative error between the two types of ratio estimations (Fig.~\ref{fig:surrogate_ratio_gap}). The rank correlation remains around $0.2$, whereas relative errors
decrease during training, falling below $5\%$ for at least $90\%$
of the sampled chunks near the end. These results show improving numerical agreement with the reference despite modest agreement in ranking.
Definitions and evaluation details are provided in
Appendix~\ref{app:ratio_gap_details}.

\noindent\textbf{Action value and diversity.} To intuitively understand how RoboFFT changes the action distribution, we visualize a representative successful rollout in Fig.~\ref{fig:viz_rollout}. We train an auxiliary critic $Q_{\mathrm{viz}}(s,a)$ to assess action quality and compare action chunks sampled from the pretrained and finetuned policies under the same trajectory states. Around key phases such as grasping, transport, and alignment, the finetuned policy shows a clear increase in $\Delta Q_{\mathrm{viz}}$ while reducing sample-based action diversity. This suggests that RoboFFT does not merely inject random exploration; it progressively concentrates the action distribution around higher-quality actions at decision-critical states. Detailed visualization protocols and a complementary t-SNE action-embedding analysis are provided in Appendix~\ref{app:visualization_details}.

\subsection{Evaluation on Real World}
\label{sec:real}

To evaluate the effectiveness of \textbf{RoboFFT} on real world robotic manipulation tasks, we integrate our method into a practical real-world RL framework~\citep{luo2024serl, luo2025precise, chen2025conrft, lei2025rl}. However, directly applying PPO-style online optimization to real robots is challenging due to limited interaction budgets, slow data collection, and the absence of parallel environments. Inspired by RL100~\citep{lei2025rl}, we therefore adapt \textbf{RoboFFT} to an offline-to-online formulation while retaining its core components including forward noising, the matching loss surrogate ratio, and PPO-style clipping. Specifically, we first pretrain a flow-based policy using human demonstrations collected via SpaceMouse teleoperation. 
We then perform multiple offline RL iterations. In each offline iteration, we perform offline PPO-style optimization on the collected rollout data, where IQL~\citep{kostrikov2021offline} is used to estimate advantages from offline trajectories. After each offline RL update, we deploy the updated policy on the real robot to collect additional trajectories, which are added to the offline dataset for the next offline RL iteration. To stabilize policy improvement and avoid drifting away from reliable human behaviors, we combine the newly collected rollout data with the original teleoperation demonstrations and conduct imitation learning to update the base policy for the next iteration.
During the online RL phase, we continue PPO-style updates initialized from the offline actor and IQL critic. We use a sparse binary $0/1$ reward judged by a human evaluator without training reward function.

\begin{figure*}[!h]
  \centering
  \vspace{-2.0mm}

      \includegraphics[width=1.0\textwidth]{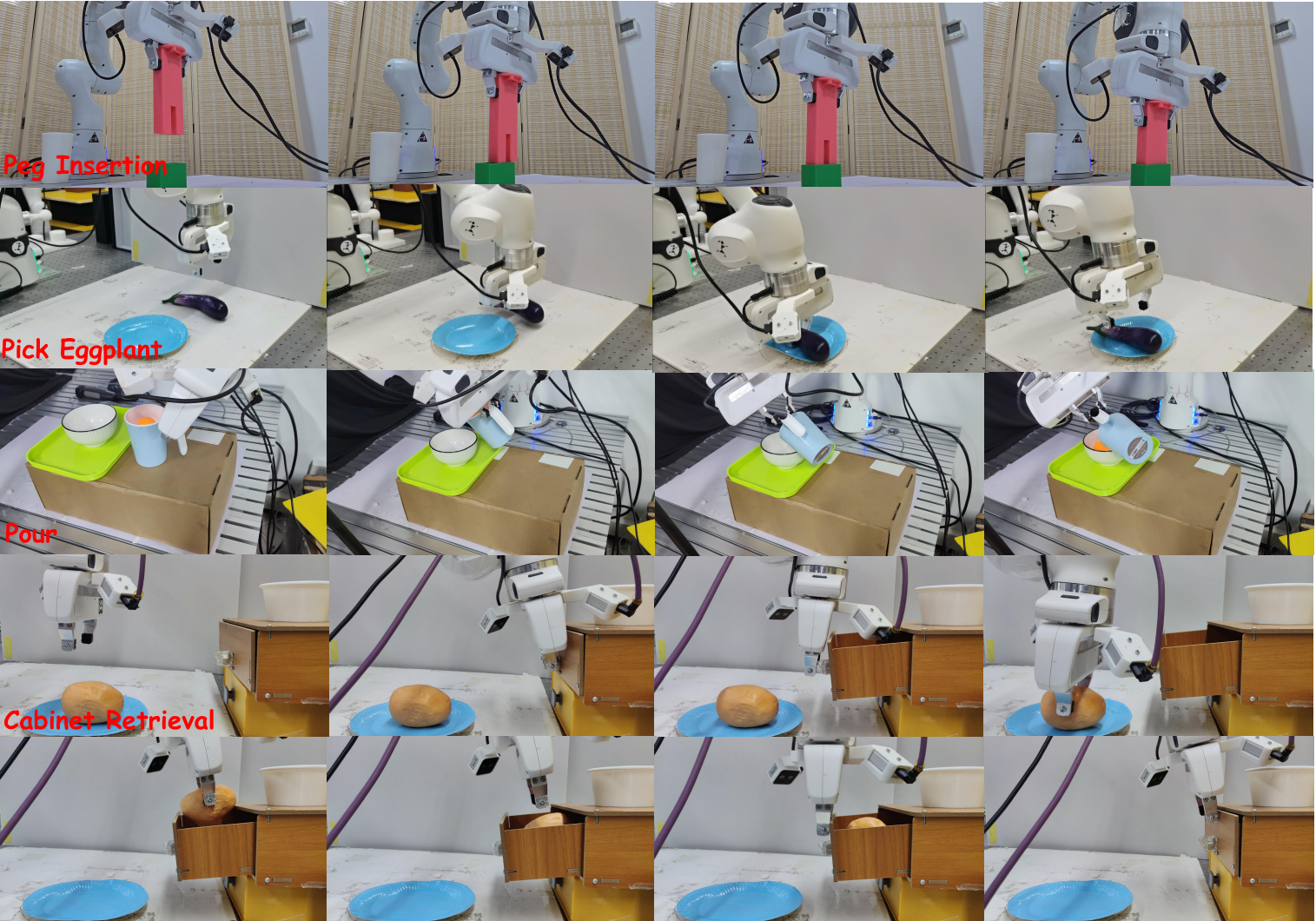}
  \caption{Successful real world rollouts on the four selected tasks..
  }

  \label{fig:real_exp_add}
  \vspace{-6mm}
  
\end{figure*}

We evaluate \textbf{RoboFFT} on four real-world tasks including \textit{Peg Insertion}, \textit{Pick Eggplant}, \textit{Pour} and \textit{Cabinet Retrieval}, covering both precision and long-horizon manipulation, as shown in Fig.~\ref{fig:real_exp_add}. For each task, we conduct 20 real world trials. 
As reported in Fig.~\ref{tab:real_world}, \textbf{RoboFFT} consistently improves performance from the pretrained policy to offline and further to online finetuning, even on \textit{Cabinet Retrieval}, where the pretrained policy has a relatively low success rate. These consistent gains demonstrate that RL post-training can effectively improve pretrained generative policies through task-level interaction feedback.
To separate the effect of RL optimization from simple data aggregation, we further compare different post-training strategies on \textit{Pick Eggplant} under the same data budget. BC Aggregation improves the pretrained policy from $12/20$ to only $14/20$, while DRWR achieves $12/20$, compared with $18/20$ for offline \textbf{RoboFFT}. Online \textbf{RoboFFT} further reaches $20/20$, matching RL100. These results show that the gains arise from effective RL post-training rather than merely adding rollout data, while online interaction provides further improvement beyond offline optimization. More implementation details are provided in Appendix~\ref{app:real_details}.

\section{Limitations}

Although \textbf{RoboFFT} enables stable online finetuning of generative robot policies through the forward noising process, it still has several limitations. First, the weighted score/flow matching loss is only a surrogate for the true action log-likelihood, and its scale may vary across noise schedules, action dimensions, and policy architectures. Therefore, additional calibration such as temperature scaling is still needed to ensure stable PPO-style updates. Second, our current framework relies on on-policy PPO, which has limited data efficiency and cannot fully reuse historical trajectories. This makes real-world deployment challenging, especially when data collection is restricted to a single robot. Although offline-to-online paradigms such as RL100~\citep{lei2025rl} can partially alleviate this issue, they usually require iterative data collection and offline retraining, which remains time-consuming in practice. Future work may explore more sample-efficient variants that better leverage historical trajectories and human intervention data for safer and faster real world robot learning.


\section{Conclusion}
\label{sec:conclusion}
We present \textbf{RoboFFT}, a forward-process reinforcement learning framework for finetuning generative robot policies. By applying forward noising to sampled actions and using the weighted score/flow matching loss to construct a surrogate policy ratio, \textbf{RoboFFT} enables PPO-style policy optimization without requiring explicit likelihood estimation. Experiments on representative simulation benchmarks show that \textbf{RoboFFT} consistently improves popular generative robot policies on long-horizon and sparse-reward manipulation tasks, while achieving better training stability and efficiency. Further integration into a real world RL framework also demonstrates its effectiveness on real robot tasks.


\clearpage
\acknowledgments{
We would like to sincerely thank Qiuxuan Chen, Yuhang Wu, and Xinyu Liu for their generous help and valuable contributions to the real world implementation and experiments. We are especially grateful to Huayu Chen (Tsinghua University) for the many insightful discussions and thoughtful suggestions at the early stage of this project. This work was supported by the National Natural Science Foundation of China (Grant No. 62561160152) and the Zhongguancun Academy (Grant No. C20250604).
}


\bibliography{references}  

\newpage

\appendix


  

\section{Additional Method Details}
\label{app:method_details}

\subsection{ELBO Interpretation of the Surrogate Ratio}
\label{app:elbo_interpretation}

RoboFFT constructs the policy ratio from the same score / flow matching loss used
for imitation pretraining. This follows the forward-process surrogate-ratio idea
used in flow policy optimization~\citep{mcallister2025flow,yi2026flow} and is
motivated by the connection between weighted denoising objectives and the
ELBO~\citep{kingma2023understanding}. For a rollout action $a_t^0$ conditioned on
$o_t$, write the conditional ELBO as
\begin{equation}
\label{eq:app_elbo}
    \mathrm{ELBO}_{\theta}(a_t^0|o_t)
    =
    \log \pi_{\theta}(a_t^0|o_t)
    -
    D_{\theta}^{\mathrm{KL}}(a_t^0,o_t),
\end{equation}
where $D_{\theta}^{\mathrm{KL}}(a_t^0,o_t)$ denotes the variational gap between
the true conditional log-likelihood and the ELBO. Taking the ratio of
exponentiated ELBOs under the current and previous policies gives
\begin{equation}
\label{eq:app_elbo_ratio}
\begin{aligned}
    \widetilde{\rho}_t(\theta)
    &=
    \frac{
        \exp\left(\mathrm{ELBO}_{\theta}(a_t^0|o_t)\right)
    }{
        \exp\left(\mathrm{ELBO}_{\theta_{\mathrm{old}}}(a_t^0|o_t)\right)
    } \\
    &=
    \frac{
        \pi_{\theta}(a_t^0|o_t)
    }{
        \pi_{\theta_{\mathrm{old}}}(a_t^0|o_t)
    }
    \exp\left(
        D_{\theta_{\mathrm{old}}}^{\mathrm{KL}}(a_t^0,o_t)
        -
        D_{\theta}^{\mathrm{KL}}(a_t^0,o_t)
    \right).
\end{aligned}
\end{equation}
Thus, the ELBO ratio can be viewed as a surrogate likelihood ratio with an
additional correction induced by the variational gap.

For common diffusion parameterizations and monotonic weightings,
\citet{kingma2023understanding} show that weighted denoising objectives are
closely related to ELBO objectives up to constants independent of the learnable
parameters. Following this connection, we use the score / flow matching loss
defined in Sec.~\ref{subsec:prelim_diffusion_and_flow} as a surrogate negative
log-likelihood:
\begin{equation}
\label{eq:app_loss_elbo}
    \mathcal{L}_{\theta}^{w}(a_t^0,o_t)
    \approx
    -
    \mathrm{ELBO}_{\theta}(a_t^0|o_t)
    +
    c(a_t^0,o_t),
\end{equation}
where $c(a_t^0,o_t)$ does not depend on $\theta$ and therefore cancels in the
ratio. Substituting Eq.~\eqref{eq:app_loss_elbo} into
Eq.~\eqref{eq:app_elbo_ratio} yields
\begin{equation}
\label{eq:app_surrogate_ratio}
    \widehat{\rho}_t(\theta)
    =
    \exp\left(
        \mathcal{L}_{\theta_{\mathrm{old}}}^{w}(a_t^0,o_t)
        -
        \mathcal{L}_{\theta}^{w}(a_t^0,o_t)
    \right),
\end{equation}
which is the unscaled surrogate ratio in Eq.~\eqref{eq:surrogate_ratio}. In
practice, we estimate the matching loss with Eq.~\eqref{eq:mc_matching_loss} and
use the temperature-scaled ratio in Eq.~\eqref{eq:scaled_ratio}.

\subsection{Advantage and Value Estimation}
\label{app:advantage_value}

For policy optimization, we estimate the advantage with generalized advantage
estimation (GAE):
\begin{equation}
\label{eq:app_gae}
    \hat{A}_t
    =
    \sum_{l=0}^{T-t}
    (\gamma\lambda_{\mathrm{GAE}})^l
    \left(
        r_{t+l}
        +
        \gamma V_{\phi}(o_{t+l+1})
        -
        V_{\phi}(o_{t+l})
    \right),
\end{equation}
where $\gamma$ is the discount factor and $\lambda_{\mathrm{GAE}}$ controls the
bias-variance trade-off. The value function is trained with the squared
regression loss
\begin{equation}
\label{eq:app_value_loss}
    \mathcal{L}_{V}(\phi)
    =
    \mathbb{E}_{(o_t,\hat{R}_t)\sim\mathcal{B}}
    \left[
        \left(
            V_{\phi}(o_t)-\hat{R}_t
        \right)^2
    \right],
\end{equation}
where $\hat{R}_t$ is the return target.

\subsection{Training Algorithm}
\label{app:algorithm}

\begin{algorithm}[tb]
\caption{RoboFFT}
\label{alg:RoboFFT}
\begin{algorithmic}[1]
\Require Pretrained prediction network $u_{\theta}$, value function $V_{\phi}$,
rollout buffer $\mathcal{B}$, number of Monte Carlo samples $N_{\mathrm{mc}}$,
ratio temperature $\alpha_{\mathrm{ratio}}$, PPO clipping threshold
$\epsilon_{\mathrm{clip}}$.
\For{each iteration}
    \State Set $\theta_{\mathrm{old}}\leftarrow\theta$.
    \State Collect online rollouts with the current generative policy; for
    flow-based policies, use the stochastic sampler in Eq.~\eqref{eq:sde_sampling}.
    \State Store transitions $(o_t,a_t^0,r_t,o_{t+1})$ in $\mathcal{B}$.
    \State Compute advantage estimates $\hat{A}_t$ and value targets $\hat{R}_t$.
    \For{each policy update epoch}
        \For{each minibatch from $\mathcal{B}$}
            \State Draw $\{(k_i,\epsilon_i)\}_{i=1}^{N_{\mathrm{mc}}}$ with
            $k_i\sim\mathcal{U}(0,1)$ and $\epsilon_i\sim\mathcal{N}(0,I)$.
            \State Construct noised actions
            $a_t^{k_i}=\alpha_{k_i}a_t^0+\sigma_{k_i}\epsilon_i$.
            \State Compute
            $\widehat{\mathcal{L}}_{\theta}^{w}(a_t^0,o_t)$ and
            $\widehat{\mathcal{L}}_{\theta_{\mathrm{old}}}^{w}(a_t^0,o_t)$
            using Eq.~\eqref{eq:mc_matching_loss}.
            \State Compute the scaled ratio
            $\widehat{\rho}^{\,s}_t(\theta)$ using Eq.~\eqref{eq:scaled_ratio}.
            \State Update $\theta$ by maximizing Eq.~\eqref{eq:robofft_objective}
            with $\widehat{\rho}_t(\theta)$ replaced by
            $\widehat{\rho}^{\,s}_t(\theta)$.
            \State Update $V_{\phi}$ by minimizing Eq.~\eqref{eq:app_value_loss}.
        \EndFor
    \EndFor
    \State Clear the rollout buffer $\mathcal{B}$.
\EndFor
\State \Return Fine-tuned generative policy $u_{\theta}$.
\end{algorithmic}
\end{algorithm}

\section{Implementation Details for the Robomimic Experiments}
\label{app:robomimic_impl}

This section provides the implementation details for the Robomimic online finetuning experiments in Sec.~\ref{sec:experiments}. We follow the Robomimic benchmark protocol~\citep{mandlekar2021matters} and use the same processed demonstration data and environment preprocessing as the DPPO and ReinFlow Robomimic experiments~\citep{ren2024diffusion,zhang2025reinflow}. All Robomimic tasks use sparse rewards and report success rates. The online interaction budget shown in the training curves is computed as
\begin{equation}
    N_{\mathrm{online}}
    =
    H_{\mathrm{exec}}\cdot n_{\mathrm{env}}\cdot n_{\mathrm{step}}\cdot n_{\mathrm{itr}},
    \label{eq:app_online_budget}
\end{equation}
where \(H_{\mathrm{exec}}\) is the number of low-level actions executed after each policy inference, \(n_{\mathrm{env}}\) is the number of parallel environments, \(n_{\mathrm{step}}\) is the number of decision steps collected per iteration, and \(n_{\mathrm{itr}}\) is the number of online finetuning iterations.

\subsection{Robomimic Benchmark}
\label{app:robomimic_benchmark}

We use Robomimic~\citep{mandlekar2021matters} as the simulated manipulation benchmark throughout our main experiments. Robomimic provides sparse-reward robot manipulation tasks collected from human demonstrations, making it a suitable testbed for evaluating whether online finetuning can improve a behavior-cloned generative policy beyond its initial demonstration-trained performance. Following the data processing and environment conventions used in DPPO and ReinFlow~\citep{ren2024diffusion,zhang2025reinflow}, we evaluate four Robomimic tasks: Lift, Can, Square, and Transport. The corresponding observation and action specifications are summarized in Table~\ref{tab:app_robomimic_specs}.

The four tasks cover different manipulation difficulties. Lift requires the robot to grasp and lift an object, while Can is a pick-and-place task where the robot must manipulate a can into the target container. Square is a more precision-sensitive assembly task in which the robot must manipulate a square nut and place it onto a positioning peg. Transport is a long-horizon bimanual manipulation task that requires coordinated control of two robot arms to move an object across the workspace. Thus, the benchmark spans both short-horizon single-arm manipulation and longer-horizon bimanual manipulation.

We consider both state-input and pixel-input variants. In the state-input setting, the policy observes low-dimensional proprioceptive and object state features. In the pixel-input setting, the policy observes RGB image(s) at \(96\times 96\) resolution together with a compact proprioceptive state. Lift, Can, and Square use a single camera view, while Transport uses two camera views. For all tasks, the action is continuous. Lift, Can, and Square use a 7-dimensional action space, whereas Transport uses a 14-dimensional action space due to its bimanual control setup.

All Robomimic tasks use sparse rewards. We report success rate as the primary evaluation metric. The maximum episode length \(T_{\max}\) is task-dependent: 300 decision steps for Lift and Can, 400 for Square, and 800 for Transport. In online finetuning, the policy predicts action chunks and executes the first \(H_{\mathrm{exec}}\) actions in each chunk; the detailed online interaction budget and algorithm-specific hyperparameters are described in the following sections.

\begin{table}[t]
\centering
\caption{Robomimic task specifications used in our online finetuning experiments. For pixel-input tasks, ``Prop. dim'' denotes the low-dimensional proprioceptive input dimension paired with image observations, and ``Image views'' reports the number and spatial resolution of RGB camera views.}
\label{tab:app_robomimic_specs}
\small
\setlength{\tabcolsep}{5.5pt}
\begin{tabular}{lcccccc}
\toprule
Task
& State obs dim
& Pixel prop. dim
& Pixel image views
& Act dim
& \(T_{\max}\)
& Reward \\
\midrule
Lift
& 19
& 9
& \(1\times 96\times 96\)
& 7
& 300
& sparse \\
Can
& 23
& 9
& \(1\times 96\times 96\)
& 7
& 300
& sparse \\
Square
& 23
& 9
& \(1\times 96\times 96\)
& 7
& 400
& sparse \\
Transport
& 59
& 18
& \(2\times 96\times 96\)
& 14
& 800
& sparse \\
\bottomrule
\end{tabular}
\end{table}

\subsection{Pretraining Setup}
\label{app:pretraining_setup}

All flow-based policies are pretrained with behavior cloning using the conditional flow matching objective, while diffusion based ones pretrained using the denoising score matching objective. The pretraining datasets follow the processed Robomimic data used by DPPO and ReinFlow, so no additional demonstrations are introduced for RoboFFT. In the state-input setting, the actor and critic are MLP-based networks. In the pixel-input setting, both the actor and critic use a ViT-based visual backbone with a proprioceptive encoder, following the architecture convention used in prior diffusion and flow policy finetuning works~\citep{ren2024diffusion,zhang2025reinflow}. For fairness, baseline policies use the same observation preprocessing, action chunking, and model sizes whenever applicable. Table~\ref{tab:app_pretrain_epochs} lists the pretrained checkpoints used as base policies in the main Robomimic experiments.

\begin{table}[t]
\centering
\caption{Base policy checkpoints used for Robomimic online finetuning. The listed numbers are the behavior-cloning pretraining epochs of the selected checkpoints.}
\label{tab:app_pretrain_epochs}
\begin{tabular}{lccc}
\toprule
Task & Action chunk \((H\times d_a)\) & State-input checkpoint & Pixel-input checkpoint \\
\midrule
Lift      & \(4\times 7\)  & 3000 epochs & 2000 epochs \\
Can       & \(4\times 7\)  & 3000 epochs & 2000 epochs \\
Square    & \(4\times 7\)  & 3000 epochs & 2000 epochs \\
Transport & \(8\times 14\) & 3000 epochs & 2000 epochs \\
\bottomrule
\end{tabular}
\end{table}

\subsection{Online Finetuning Budget and Shared Environment Configuration}
\label{app:budget_env_config}

All methods are evaluated under the same Robomimic environment wrappers and action chunking convention. A policy receives one observation step, i.e., \(\texttt{cond\_steps}=1\), predicts an action chunk of horizon \(H\), and executes the first \(H_{\mathrm{exec}}\) low-level actions. For Lift, Can, and Square, we use \(H=H_{\mathrm{exec}}=4\); for Transport, we use \(H=H_{\mathrm{exec}}=8\). The episode caps are 300 decision steps for Lift and Can, 400 for Square, and 800 for Transport. We keep these task and model specifications aligned across RoboFFT and all baselines; the curves are plotted against the online sample budget in Eq.~\eqref{eq:app_online_budget}.

\begin{table}[t]
\centering
\caption{Online finetuning budgets for Robomimic experiments. The last column equals \(H_{\mathrm{exec}}\cdot n_{\mathrm{env}}\cdot n_{\mathrm{step}}\cdot n_{\mathrm{itr}}\), which gives the total number of online interaction samples.}
\label{tab:app_online_budget}
\begin{tabular}{llccccc}
\toprule
Task & Input & \(H_{\mathrm{exec}}\) & \(n_{\mathrm{env}}\) & \(n_{\mathrm{step}}\) & \(n_{\mathrm{itr}}\) & Online samples \\
\midrule
Lift      & state & 4 & 50 & 300 & 81  & 4.86M \\
Lift      & pixel & 4 & 50 & 300 & 81  & 4.86M \\
Can       & state & 4 & 50 & 300 & 151 & 9.06M \\
Can       & pixel & 4 & 50 & 300 & 151 & 9.06M \\
Square    & state & 4 & 50 & 400 & 201 & 16.08M \\
Square    & pixel & 4 & 50 & 400 & 201 & 16.08M \\
Transport & state & 8 & 50 & 400 & 201 & 32.16M \\
Transport & pixel & 8 & 50 & 400 & 201 & 32.16M \\
\bottomrule
\end{tabular}
\end{table}

\subsection{Baseline Methods and Configurations}
\label{app:baselines}

We compare RoboFFT with representative online RL fine-tuning baselines for diffusion- and flow-based robot policies.

\textbf{DPPO.}
Diffusion Policy Policy Optimization~\citep{ren2024diffusion} is a representative reverse-process fine-tuning method for diffusion policies. It formulates the reverse denoising process together with the environment interaction process as a two-level MDP, and applies PPO to optimize the Gaussian likelihoods of denoising transitions. We use DPPO as the canonical diffusion-policy RL baseline for long-horizon manipulation tasks.

\textbf{ReinFlow.}
ReinFlow~\citep{zhang2025reinflow} extends the reverse-process policy-gradient paradigm to flow-based policies. It injects stochasticity into the deterministic flow trajectory, turns the sampler into a discrete-time Markov process with tractable Gaussian transition probabilities, and performs PPO-style optimization over the internal sampling trajectory. We use ReinFlow as the primary flow-policy baseline.

\textbf{FPO.}
Flow Policy Optimization~\citep{mcallister2025flow} is a forward-process policy-gradient method that constructs a PPO-style surrogate ratio from conditional flow matching losses. We adapt FPO into our online fine-tuning codebase for Robomimic. To ensure fair comparison under the same tasks and pretrained policies, most environment and optimizer settings are shared with RoboFFT. The main algorithmic differences are that FPO does not use our SDE explorative sampling strategy and ratio-temperature calibration adopted in RoboFFT, while following the original FPO convention of using $N_{\mathrm{mc}}=8$ $(k_i,\epsilon_i)$ samples for ratio estimation.

\textbf{DRWR.}
DRWR is a lightweight diffusion reward-weighted regression baseline following the implementation convention in DPPO~\citep{ren2024diffusion}. It performs forward-noise matching style updates weighted by rewards or returns, and serves as a simple forward-process reference that is architecturally close to our approach but does not use the PPO-style objective and the many critical designs introduced by our work.

For all baselines, we match the Robomimic environment wrappers, observation preprocessing, action chunking, base-policy training data, and evaluation protocol whenever applicable. The hyperparameters of DPPO, ReinFlow, and DRWR follow their original implementations unless a task-specific adaptation is necessary. Since the released ReinFlow manipulation implementation focuses on pixel-input Robomimic tasks, for state-input tasks we adapt secondary settings such as optimizer and update-batch configurations while keeping the core ReinFlow design unchanged. For fair comparison, the number of inference steps and finetuning steps are tuned on each task, so that each baseline achieves its nearly best performance. RoboFFT-F and RoboFFT-D are respectively implemented upon the ReinFlow ~\citep{zhang2025reinflow} and DPPO ~\citep{ren2024diffusion} codebases, and use the same online interaction budget as the corresponding baselines on each task. We report the task success rate (fraction of multiple parallelized rollouts that complete the task at each evaluation point) as the principal performance metric for Robomimic tasks.


\subsection{RoboFFT-F Hyperparameters}
\label{app:robofft_f_hparams}

Unless stated otherwise, RoboFFT uses 10 sampling steps during both rollout and evaluation and sets \(\texttt{ft\_denoising\_steps}=10\) during finetuning. Exploration is enabled by the stochastic forward-process perturbation in Eq.~\eqref{eq:sde_sampling}; in all main Robomimic runs we set the fixed denoising noise level to \(\eta=0.25\), implemented by \(\texttt{min\_std}=\texttt{max\_std}=0.25\). We use \(\texttt{sample\_t\_type}=\texttt{linspace}\), i.e., the flow time used to compute the CFM surrogate loss is sampled uniformly from the discrete integration grid used at inference. The policy loss clipping coefficient is \(\texttt{clip\_ploss\_coef}=0.02\). 

In Eq.~\eqref{eq:scaled_ratio}, we calibrate the CFM-loss difference by \(\alpha_{\mathrm{ratio}}\). In the paper-level notation, the surrogate log-ratio is normalized by the action feature dimension, but not by the action horizon. Under this convention, we set \(\alpha_{\mathrm{ratio}}=2\) for Lift, Can, and Square, and \(\alpha_{\mathrm{ratio}}=1\) for Transport. The smaller value for Transport compensates for its longer action chunk horizon. The Monte-Carlo repetition number is \(N_{\mathrm{mc}}=1\) for all main runs except Square-state, where we use \(N_{\mathrm{mc}}=2\). Table~\ref{tab:app_robofft_hparams} summarizes the task-specific optimizer settings.

\begin{table}[t]
\centering
\caption{Main RoboFFT-F optimizer hyperparameters for Robomimic. All runs use \(\gamma=0.999\), GAE parameter \(\lambda=0.95\), 10 update epochs per iteration, target KL \(10^{-2}\), and the repeat-samples trick described in Sec.~\ref{subsec:practical_implementation}.}
\label{tab:app_robofft_hparams}
\begin{tabular}{llccccc}
\toprule
Task & Input & Actor LR & Critic LR & Batch size & \(N_{\mathrm{mc}}\) & \(\alpha_{\mathrm{ratio}}\) \\
\midrule
Lift      & state & \(2\times10^{-6}\)   & \(1\times10^{-3}\)    & 10000 & 1 & 2 \\
Can       & state & \(2\times10^{-6}\)   & \(1\times10^{-3}\)    & 10000 & 1 & 2 \\
Square    & state & \(2\times10^{-5}\)   & \(1\times10^{-3}\)    & 10000 & 2 & 2 \\
Transport & state & \(2\times10^{-6}\)   & \(1\times10^{-3}\)    & 10000 & 1 & 1 \\
\midrule
Lift      & pixel & \(2\times10^{-5}\)   & \(6.5\times10^{-4}\)  & 500   & 1 & 2 \\
Can       & pixel & \(2\times10^{-5}\)   & \(6.5\times10^{-4}\)  & 500   & 1 & 2 \\
Square    & pixel & \(3.5\times10^{-6}\) & \(4.5\times10^{-4}\)  & 500   & 1 & 2 \\
Transport & pixel & \(3.5\times10^{-6}\) & \(4.5\times10^{-4}\)  & 500   & 1 & 1 \\
\bottomrule
\end{tabular}
\end{table}

\subsection{Specialized Designs for RoboFFT-D}
\label{app:robofft_d_designs}

RoboFFT-D is the diffusion-policy instantiation of RoboFFT and is implemented upon the DPPO codebase~\citep{ren2024diffusion}. We keep the Robomimic environment wrappers, diffusion actor / critic architectures, rollout buffer, GAE estimation, PPO update loop, and evaluation protocol aligned with DPPO whenever applicable. The key difference is that RoboFFT-D does not treat the reverse denoising chain as a diffusion MDP for computing Gaussian transition likelihoods. Instead, it applies forward noising to the final rollout action \(a_t^0\) and uses a weighted denoising score matching loss \(\mathcal{L}_{\theta}^{w,D}(a_t^0,o_t)\) as the surrogate negative log-likelihood for PPO-style policy optimization.

\noindent\textbf{Delta-log-SNR weighting.}
For RoboFFT-F, we use the simplest flow matching weighting $w(\lambda_k)$ so that the effective coefficient before the score / flow matching MSE is constant: $\Omega_F(k)=w(\lambda_k)(-\mathrm{d}\lambda_k/\mathrm{d}k)\equiv 1$. This makes the CFM surrogate loss reduce to the standard unweighted velocity-matching MSE loss used in flow pretraining. RoboFFT-D instead uses a diffusion-specific discrete log-SNR weighting. Let \(\bar{\alpha}_j\) denote the cumulative product of diffusion alphas at denoising index \(j\). We define a clipped variance and the corresponding clipped log-SNR by
\begin{equation}
\label{eq:app_tilde_lambda}
\begin{aligned}
    \widetilde{\sigma}_j^2
    &=
    \max\left(
        1-\bar{\alpha}_j,\,
        \sigma_{\min}^2
    \right), \\
    \widetilde{\lambda}_j
    &=
    \log
    \frac{
        \max\left(1-\widetilde{\sigma}_j^2,\,\sigma_{\min}^2\right)
    }{
        \widetilde{\sigma}_j^2
    } .
\end{aligned}
\end{equation}
For a sampled denoising index \(j\), the discrete weighting coefficient is
\begin{equation}
\label{eq:app_delta_tilde_lambda}
    \widetilde{\Delta\lambda}_j
    =
    \widetilde{\lambda}_{j-1}
    -
    \widetilde{\lambda}_{j},
\end{equation}
where the previous index follows the corresponding DDPM or DDIM finetuning grid. This discrete coefficient approximates \((-\mathrm{d}\lambda_k/\mathrm{d}k)\) and the weighting schedule is chosen as $w(\lambda_k)\equiv1$. It keeps the surrogate matching loss closer to the weighted denoising objective motivated by the ELBO interpretation~\citep{kingma2023understanding}, while the clipping in \(\widetilde{\lambda}_j\) avoids excessively large weights near nearly clean actions.

\noindent\textbf{Diffusion surrogate matching loss.}
For each rollout action \(a_t^0\), RoboFFT-D samples \(N_{\mathrm{mc}}\) forward-noising pairs and evaluates the noise-prediction error of the diffusion policy \(u_{\theta}=\epsilon_{\theta}\). Specifically, for the \(i\)-th sample,
\begin{equation}
\label{eq:app_robofft_d_noising}
    a_t^{k_i}
    =
    \alpha_{k_i}a_t^0+\sigma_{k_i}\epsilon_i,
    \qquad
    \epsilon_i\sim\mathcal{N}(0,I).
\end{equation}
The Monte Carlo estimate of the diffusion-specific weighted score matching loss is
\begin{equation}
\label{eq:app_robofft_d_loss}
\begin{aligned}
    \widehat{\mathcal{L}}_{\theta}^{w,D}(a_t^0,o_t)
    =
    \frac{1}{N_{\mathrm{mc}}}
    \sum_{i=1}^{N_{\mathrm{mc}}}
    \sum_{h=1}^{H}
    \frac{1}{d_a}
    \sum_{d=1}^{d_a}
    \frac{\widetilde{\Delta\lambda}_{j_i}}{2}
    \left(
        \epsilon_{\theta}(a_t^{k_i},k_i,o_t)_{h,d}
        -
        \epsilon_{i,h,d}
    \right)^2 .
\end{aligned}
\end{equation}
Here \(H\) is the action horizon and \(d_a\) is the action dimension. The implementation normalizes the score matching error by the action dimension but not by the action horizon, which is consistent with the implementation of RoboFFT-F. The resulting diffusion-specific surrogate ratio is
\begin{equation}
\label{eq:app_robofft_d_ratio}
    \widehat{\rho}^{D}_t(\theta)
    =
    \exp\left(
        \alpha_{\mathrm{ratio}}
        \left[
            \widehat{\mathcal{L}}_{\theta_{\mathrm{old}}}^{w,D}(a_t^0,o_t)
            -
            \widehat{\mathcal{L}}_{\theta}^{w,D}(a_t^0,o_t)
        \right]
    \right),
\end{equation}
which follows the same matching-loss-ratio form as Eq.~\eqref{eq:surrogate_ratio} and is used in the clipped PPO-style objective in Eq.~\eqref{eq:robofft_objective}.

\noindent\textbf{High-repetition Monte Carlo estimation.}
RoboFFT-D uses a notably larger number of Monte Carlo forward-noising samples than RoboFFT-F. In the main diffusion-policy Robomimic runs, we use $N_{\mathrm{mc}}=16$. This is important because the denoising score matching loss can be noisier than the flow matching loss when used as a surrogate negative log-likelihood, since the denoising process is intrinstically stochastic, and RoboFFT-D additionally enforces a lowerbound $0.1$ on the single step sampling variance $\sigma_j$ at sampling time as in DPPO, which aims for better exploration and induces more stochasticity to the score matching loss. A small number of forward-noising samples may produce a high-variance estimate of the matching-loss difference, especially when sampled timesteps fall near high-curvature regions of the log-SNR schedule. Averaging over multiple samples reduces the variance of the surrogate ratio and prevents the PPO update from being dominated by accidental high-error denoising samples. 

\noindent\textbf{Best-evaluation anchor regularization.}
For difficult manipulation tasks, RoboFFT-D additionally uses a best-evaluation anchor network. We maintain a frozen anchor policy \(\pi_{\bar{\theta}}\), initialized from the pretrained diffusion policy. After each evaluation phase, if the current policy achieves a success rate no worse than the best recorded evaluation success rate, we update the anchor parameters:
\begin{equation}
\label{eq:app_anchor_update}
    \bar{\theta}
    \leftarrow
    \theta,\ \mathrm{SR}_{\mathrm{best}} \gets \mathrm{SR}_{\mathrm{eval}}(\theta)
    \quad
    \text{if}
    \quad
    \mathrm{SR}_{\mathrm{eval}}(\theta)
    \geq
    \mathrm{SR}_{\mathrm{best}} .
\end{equation}
The anchor loss then regularizes the trainable policy toward actions sampled from this best-evaluation policy:
\begin{equation}
\label{eq:app_anchor_loss}
\begin{aligned}
    a^{0,\mathrm{anc}}_t
    &\sim
    \pi_{\bar{\theta}}(\cdot|o_t), \\
    \mathcal{L}_{\mathrm{anchor}}(\theta)
    &=
    \mathbb{E}_{o_t,\,a^{0,\mathrm{anc}}_t}
    \left[
        \widehat{\mathcal{L}}_{\theta}^{w,D}
        (a^{0,\mathrm{anc}}_t,o_t)
    \right].
\end{aligned}
\end{equation}
Equivalently, this applies a BC-style denoising score matching loss to samples generated by the best historical policy. It helps prevent the diffusion policy from drifting away from empirically successful action regions. When enabled, the full actor loss is
\begin{equation}
\label{eq:app_robofft_d_actor_loss}
    \mathcal{L}_{\mathrm{actor}}^{D}
    =
    -
    J_{\mathrm{RoboFFT-D}}(\theta)
    +
    \lambda_{\mathrm{anchor}}
    \mathcal{L}_{\mathrm{anchor}}(\theta).
\end{equation}

\noindent\textbf{Negative-advantage discount.}
RoboFFT-D also uses a task-dependent negative-advantage discount. In the diffusion score matching surrogate, samples with negative advantages can induce disproportionately large matching-loss-ratio gradients and over-penalize sampled actions, especially in sparse-reward manipulation tasks where advantage estimates are noisy. We therefore attenuate the PPO-style loss for negative-advantage samples:
\begin{equation}
\label{eq:app_negdisc_loss}
    \psi_{\mathrm{negdisc}}
    \left(
        \widehat{\rho}^{D}_t,
        \hat{A}_t
    \right)
    =
    \begin{cases}
    \psi_{\mathrm{PPO}}
    \left(
        \widehat{\rho}^{D}_t,
        \hat{A}_t
    \right),
    &
    \hat{A}_t \geq 0, \\[0.4em]
    \beta_{\mathrm{neg}}\,
    \psi_{\mathrm{PPO}}
    \left(
        \widehat{\rho}^{D}_t,
        \hat{A}_t
    \right),
    &
    \hat{A}_t < 0 ,
    \end{cases}
\end{equation}
where
\begin{equation}
\label{eq:app_ppo_element}
    \psi_{\mathrm{PPO}}
    \left(
        \rho,
        A
    \right)
    =
    \min
    \left(
        \rho A,\,
        \mathrm{clip}
        \left(
            \rho,
            1-\epsilon_{\mathrm{clip}},
            1+\epsilon_{\mathrm{clip}}
        \right)
        A
    \right).
\end{equation}
Here \(\beta_{\mathrm{neg}}\in(0,1]\) controls the strength of negative-advantage attenuation. When \(\beta_{\mathrm{neg}}=1\), the objective reduces to the standard clipped PPO-style update. This is related in spirit to the asymmetric stabilization used in FPO++~\citep{yi2026flow}, but RoboFFT-D applies a simpler direct attenuation of negative-advantage gradients, which we find robust for diffusion-policy finetuning in manipulation tasks.

\subsection{RoboFFT-D Hyperparameters}
\label{app:robofft_d_hparams}

Table~\ref{tab:app_robofft_d_hparams} summarizes the main RoboFFT-D hyperparameters used in the Robomimic experiments. Unless otherwise stated, all RoboFFT-D runs use \(\gamma=0.999\), GAE parameter \(\lambda=0.95\), target KL \(0.05\), and \(N_{\mathrm{mc}}=16\). State-input tasks use \(20\) diffusion denoising steps and finetune the last \(10\) steps, while pixel-input tasks use \(100\) diffusion denoising steps and finetune the last \(5\) steps. We use DDPM-style evaluation for state-input runs and DDIM-style evaluation for pixel-input runs, consistent with the convention of DPPO.

\begin{table}[t]
\centering
\small
\caption{Main RoboFFT-D hyperparameters for Robomimic. \(N_{\mathrm{mc}}\) denotes the number of forward-noising Monte Carlo samples used for estimating the surrogate matching loss. ``Anchor'' reports the coefficient \(\lambda_{\mathrm{anchor}}\); a dash means the anchor loss is disabled. \(\beta_{\mathrm{neg}}\) is the negative-advantage discount coefficient in Eq.~\eqref{eq:app_negdisc_loss}.}
\label{tab:app_robofft_d_hparams}
\resizebox{\textwidth}{!}{%
\begin{tabular}{llcccccccc}
\toprule
Task & Input
& Denoise steps
& FT steps
& Eval
& Actor LR
& Batch
& \(N_{\mathrm{mc}}\)
& \(\alpha_{\mathrm{ratio}}\)
& Anchor / \(\beta_{\mathrm{neg}}\)
\\
\midrule
Lift
& state
& 20
& 10
& DDPM
& \(2\times10^{-5}\)
& 2500
& 16
& 1.0
& -- / 0.125
\\
Can
& state
& 20
& 10
& DDPM
& \(2\times10^{-5}\)
& 2500
& 16
& 1.0
& -- / 0.125
\\
Square
& state
& 20
& 10
& DDPM
& \(2\times10^{-5}\)
& 2500
& 16
& 0.5
& 0.05 / 0.125
\\
Transport
& state
& 20
& 10
& DDPM
& \(2\times10^{-5}\)
& 2500
& 16
& 0.5
& 0.05 / 1.0
\\
\midrule
Lift
& pixel
& 100
& 5
& DDIM
& \(2\times10^{-5}\)
& 500
& 16
& 1.0
& -- / 0.125
\\
Can
& pixel
& 100
& 5
& DDIM
& \(2\times10^{-5}\)
& 500
& 16
& 1.0
& -- / 0.125
\\
Square
& pixel
& 100
& 5
& DDIM
& \(4\times10^{-6}\)
& 500
& 16
& 1.0
& 0.05 / 1.0
\\
Transport
& pixel
& 100
& 5
& DDIM
& \(4\times10^{-6}\)
& 500
& 16
& 0.5
& 0.05 / 1.0
\\
\bottomrule
\end{tabular}
}
\end{table}

The anchor loss is enabled mainly on the more difficult Square and Transport tasks, where diffusion-policy updates are more prone to collapse under sparse rewards and long-horizon credit assignment. The negative-advantage discount is task-dependent. Smaller values of \(\beta_{\mathrm{neg}}\) more strongly suppress negative-advantage updates, while \(\beta_{\mathrm{neg}}=1\) disables the attenuation and recovers the standard clipped PPO-style loss.

\subsection{Comparison Between RoboFFT-F and RoboFFT-D Implementations}
\label{app:robofft_f_d_comparison}

RoboFFT-F and RoboFFT-D share the same high-level forward-process policy-gradient principle: both apply forward noising to rollout actions and construct a PPO-style surrogate ratio from the corresponding score / flow matching loss. Their practical implementations differ because velocity-matching flow policies and noise-prediction diffusion policies have different sampling behavior, loss geometry, and surrogate-ratio variance.

\begin{table}[t]
\centering
\small
\caption{Implementation differences between RoboFFT-F and RoboFFT-D. Both methods use the same high-level RoboFFT objective, but require different stabilization choices due to their different generative parameterizations.}
\label{tab:app_robofft_f_d_comparison}
\resizebox{\textwidth}{!}{%
\begin{tabular}{lll}
\toprule
Component & RoboFFT-F & RoboFFT-D \\
\midrule
Base codebase
& ReinFlow-based flow policy implementation
& DPPO-based diffusion policy implementation
\\
Prediction target
& Velocity field \(u_{\theta}=v_{\theta}\)
& Noise prediction \(u_{\theta}=\epsilon_{\theta}\)
\\
Forward-process loss
& Conditional flow matching loss
& Denoising score matching loss
\\
Effective weighting
& \(\Omega_F(k)\equiv 1\)
& Discrete \(\widetilde{\Delta\lambda}\) weighting
\\
Rollout stochasticity
& Explicit SDE exploration for deterministic flow sampling
& Diffusion sampler stochasticity
\\
MC repetition
& Small \(N_{\mathrm{mc}}\), typically \(1\) or \(2\)
& Larger \(N_{\mathrm{mc}}=16\)
\\
Main stabilizers
& SDE exploration, ratio-temperature scaling
& \(\widetilde{\Delta\lambda}\) weighting, anchor loss, negative-advantage discount
\\
Hard-task regularization
& Mainly through ratio scale calibration
& anchor regularization and attenuated negative-advantage updates
\\
\bottomrule
\end{tabular}
}
\end{table}

In summary, RoboFFT-F uses a velocity-matching CFM surrogate loss with a unit effective MSE coefficient and relies on SDE exploration and ratio-temperature calibration as its main stabilization mechanisms. RoboFFT-D uses a noise-prediction DSM surrogate loss with the \(\widetilde{\Delta\lambda}\) weighting, larger Monte Carlo repetition, and, on harder tasks, anchor regularization and negative-advantage attenuation. These diffusion-specific designs are necessary because the score matching surrogate loss can have higher estimation variance and more severe negative-advantage amplification than the flow matching surrogate loss used in RoboFFT-F.

\section{Additional Analyses for the Robomimic Experiments}

\subsection{Performance Analysis}
\label{app:robomimic_additional_analysis}

This subsection provides additional discussion for the main Robomimic comparison in Sec.~\ref{sec:baseline}, where the comprehensive performance of RoboFFT-F, RoboFFT-D and the baseline methods are evaluated under both state-input and pixel-input settings.

Across the easier \textit{Lift} and \textit{Can} tasks, RoboFFT-D and RoboFFT-F reach final success rates comparable to reverse-process baselines such as DPPO and ReinFlow. Their convergence can be slightly slower in some early-stage curves, but the final policy quality remains competitive. This suggests that the forward-process surrogate ratio preserves enough policy-gradient signal for improving already competent pretrained policies, while avoiding explicit likelihood computation through the reverse denoising chain.

The advantage of RoboFFT becomes more pronounced on the harder tasks. On \textit{Square} and \textit{Transport}, especially with pixel observations, RoboFFT shows a more sustained improvement trend and stronger final performance. Reverse-process methods can suffer from instability or later performance recession, likely because optimizing through a long internal sampling chain introduces a more complicated credit-assignment path and a heavier computation graph, making the gradients and parameter updates instable. In comparison, RoboFFT directly applies forward noising to the final rollout action and optimizes a matching-loss-based surrogate ratio, which keeps the optimization path closer to imitation pretraining.

The forward-process baselines, DRWR and FPO, are generally less competitive in the Robomimic manipulation setting. DRWR uses reward-weighted regression and does not include a PPO-style trust region, making it more sensitive to high-variance reward weights in sparse-reward tasks. FPO uses a forward-process PPO-style ratio, but in our implementation it lacks the manipulation-oriented stabilization designs used in RoboFFT, including controlled SDE exploration and ratio-temperature calibration. These differences help explain why simply applying a forward matching loss is not sufficient for robust online fine-tuning on long-horizon manipulation tasks.

\subsection{Evaluation with Additional Random Seeds}
\label{app:additional_seeds}

\begin{figure*}[t]
  \centering
  \subfigure[Square-state: 10 seeds]{%
    \includegraphics[width=0.47\textwidth]
      {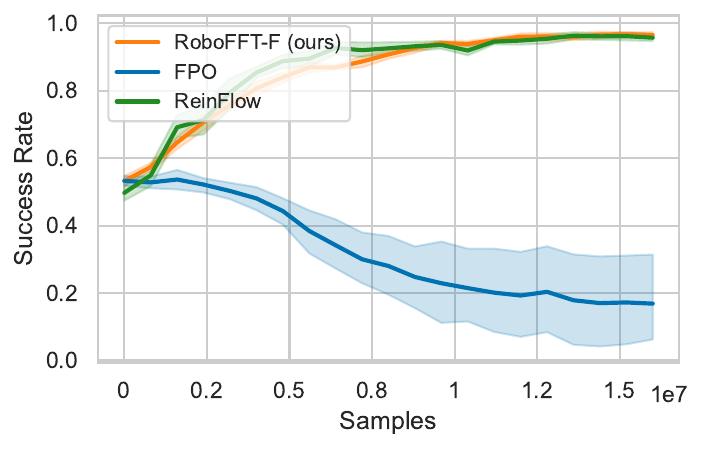}
  }
  \hfill
  \subfigure[Transport-state: 5 seeds]{%
    \includegraphics[width=0.47\textwidth]
      {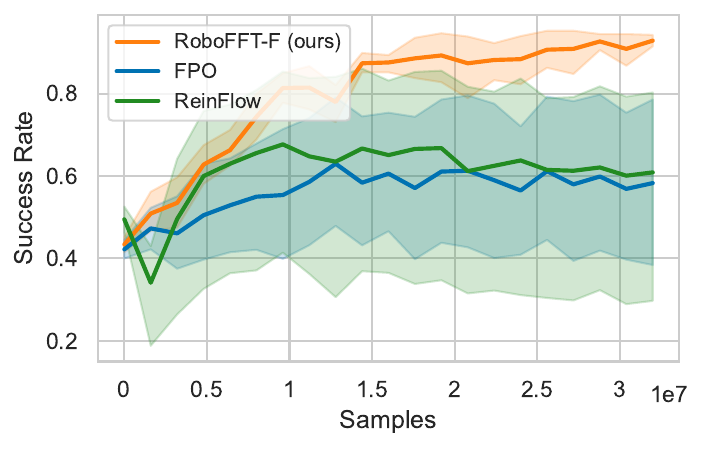}
  }
  \caption{
  Additional-seed evaluation of RoboFFT-F, FPO, and ReinFlow.
  Curves indicate means and shaded regions indicate one standard
  deviation across random seeds.
  }
  \label{fig:app_additional_seeds}
\end{figure*}

To assess sensitivity to random initialization beyond the three-seed
benchmark results in Fig.~\ref{fig:robomimic}, we evaluate RoboFFT-F,
FPO, and ReinFlow with 10 random seeds on Square-state and 5 on
Transport-state.
Figure~\ref{fig:app_additional_seeds} reports the mean success rate
and one standard deviation across seeds.

The expanded evaluations are consistent with the main results in Fig.~\ref{fig:robomimic}.
On Square-state, RoboFFT-F and ReinFlow attain similarly high success
rates, while FPO deteriorates during finetuning. On Transport-state, RoboFFT-F maintains a higher mean success rate than the two flow-policy baselines.
These results strengthen the evidence for the reported trends in Sec.~\ref{sec:baseline}.


\subsection{Efficiency Analysis}
\label{app:efficiency_analysis}

We measure the wall-clock training time per iteration for RoboFFT-F
and ReinFlow, and for RoboFFT-D and DPPO.
The training timer covers policy and value-network optimization,
excluding action sampling and environment rollout.
For each configuration, we report the mean and standard deviation
over the first 10 training iterations.
The error bars therefore characterize variation across iterations,
rather than across independent random seeds.

The comparisons are made within each policy family using the
corresponding Robomimic implementations.
For the diffusion timing runs, we disable critic-only warmup and
learning-rate warmup for both methods
(\texttt{n\_critic\_warmup\_itr}=0 and \texttt{warmup\_steps}=0).
Other algorithm-specific settings are retained, including
$N_{\mathrm{mc}}=16$ for RoboFFT-D.


For task configuration $\tau$, let
$\overline{t}_{\mathrm{base},\tau}$ and
$\overline{t}_{\mathrm{ours},\tau}$ denote the mean training times
of the baseline and RoboFFT, respectively.
We summarize efficiency using the arithmetic mean of task-wise
speedups:
\[
  S =
  \frac{1}{|\mathcal{T}|}
  \sum_{\tau\in\mathcal{T}}
  \frac{\overline{t}_{\mathrm{base},\tau}}
       {\overline{t}_{\mathrm{ours},\tau}},
\]
where $\mathcal{T}$ contains the task and observation configurations
included in the corresponding comparison.


RoboFFT-F achieves an average training-time speedup of $1.53\times$
over ReinFlow.
On Lift-state, Can-state, Square-state, and Transport-state,
RoboFFT-D achieves an average speedup of $38.0\times$ over DPPO.
The latter result shows that the cost of repeated forward-noising
evaluations does not outweigh the computational savings of the
matching-loss update in the measured diffusion implementations.

The two comparisons evaluate the complete training stages of the
respective implementations.
Their speedups can depend on network architecture, update settings,
and the number of internal sampling steps; they should not be
interpreted as an isolated measurement of a single operation.
Moreover, a training-time speedup does not translate directly into
the same full-iteration speedup, because rollout collection and
other operations contribute additional cost.

\subsection{Surrogate-Ratio Evaluation Protocol}
\label{app:ratio_gap_details}

We evaluate the matching-loss surrogate during RoboFFT-F finetuning
on Robomimic Square-state.
At each evaluated iteration, we randomly sample 2,048 action chunks to compare
the temperature-scaled surrogate in Eq.~\eqref{eq:scaled_ratio}
with a ReinFlow-style closed-form reference as is implemented by ~\cite{zhang2025reinflow}. Both ratios are evaluated for the same current and old policy
parameters using the same sampled actions.


Let $\widehat{\rho}_{j,i}$ and $\rho^{\mathrm{ref}}_{j,i}$ denote
the surrogate and reference for chunk $i$ at iteration $j$.
We compute the Spearman correlation between their sample rankings within each iteration, together with quantiles of the relative error:

\[
  e_{j,i}
  =
  \frac{
    \left|\widehat{\rho}_{j,i}-\rho^{\mathrm{ref}}_{j,i}\right|
  }{
    \left|\rho^{\mathrm{ref}}_{j,i}\right|
  }.
\]

The diagnostics in Fig.~\ref{fig:surrogate_ratio_gap} show distinct
aspects of agreement.
Spearman correlation remains around $0.2$, indicating limited
agreement in sample ordering.
Meanwhile, relative-error quantiles decrease, with errors below
$5\%$ for at least $90\%$ of samples near the end of training.
Small ratio errors do not imply high rank correlation, particularly
when reference ratios are concentrated near one.
The experiment therefore characterizes numerical discrepancy in
the evaluated setting, without establishing strict correspondence of different forms of surrogate likelihoods.

\section{Additional Ablation Studies}
\label{app:ablation_details}

\subsection{Additional RoboFFT-F Ablations on Square-State}
\label{app:ablation_logratio_alpha}

The ablation studies for RoboFFT-F in Sec.~\ref{sec:abl_viz} are conducted on Robomimic Square with state input. This task has a representative difficulty level: it is sparse-reward, requires precise object manipulation, and is sensitive to unstable online updates. Unless otherwise specified, all ablations are based on the same hyperparameters as the main configuration in Table~\ref{tab:app_robofft_hparams}. The online interaction budget and optimization schedule are kept unchanged across variants, so that each ablation isolates the intended design factor. We introduce additional ablation studies for other factors that contribute to the performance and stability of RoboFFT-F.

\noindent\textbf{Temperature Scaling in the Surrogate Ratio.} As discussed in Sec.~\ref{subsec:practical_implementation}, the score / flow matching loss is used as a surrogate negative log-likelihood through the ELBO interpretation of diffusion and flow objectives~\citep{kingma2023understanding,mcallister2025flow}. However, the numerical scale of a matching-loss difference does not necessarily match the log-scale of the true action likelihood ratio used in PPO~\citep{schulman2017proximal}. If the scale is too small, the effective policy ratio remains close to one and learning becomes slow. If the scale is too large, the ratio may frequently saturate the clipping interval and destabilize updates.

\begin{figure*}[!t]
  \centering
  \makebox[1.0\textwidth][c]{%
    \subfigure[Ratio temperature]{%
      \includegraphics[width=0.32\textwidth]{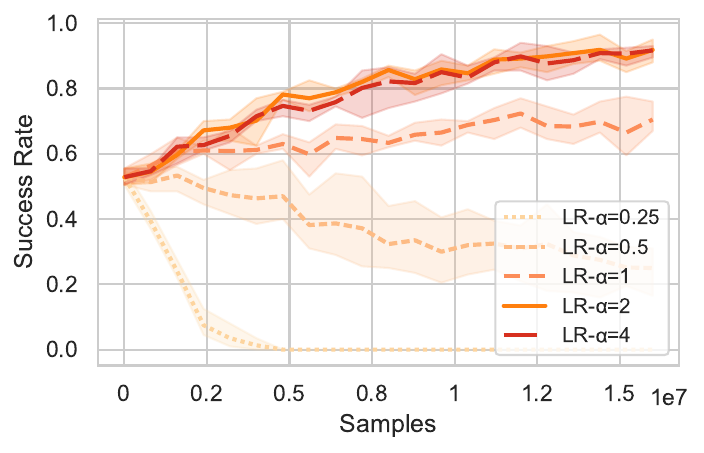}
      \label{fig:app_logratio_alpha}
    }%
    \hfill
    \subfigure[Advantage weighting]{%
      \includegraphics[width=0.32\textwidth]{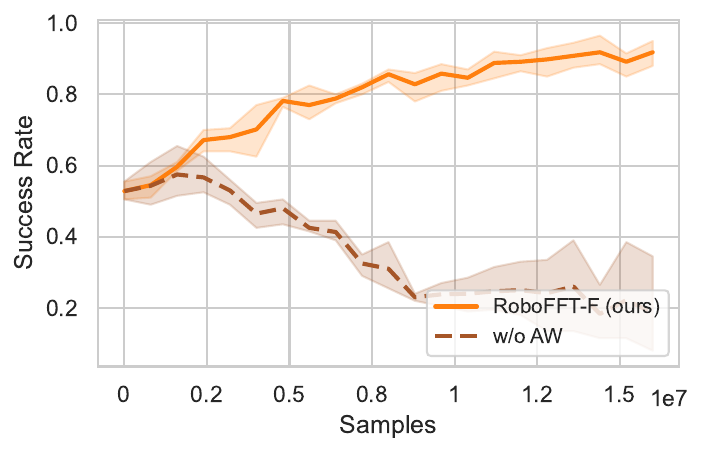}
      \label{fig:app_ablation_aw}
    }%
    \hfill
    \subfigure[PPO-style update]{%
      \includegraphics[width=0.32\textwidth]{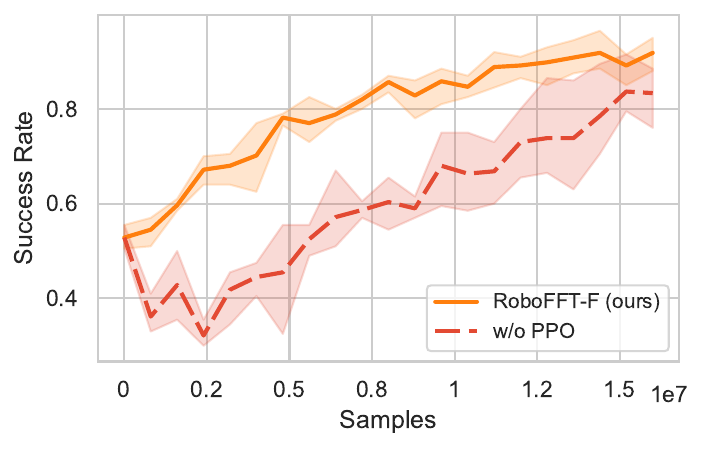}
      \label{fig:app_ablation_ppo}
    }%
  }
  \caption{
  Additional ablation studies on Robomimic Square-state. We evaluate the effects of ratio temperature, advantage weighting, and PPO-style clipping. Ratio temperature calibrates the surrogate likelihood-ratio scale, advantage weighting reduces unstable return-weighted updates, and PPO-style clipping stabilizes early finetuning by constraining the surrogate-ratio update.
  }
  \label{fig:app_additional_ablations}
\end{figure*}

We therefore introduce the ratio-temperature coefficient \(\alpha_{\mathrm{ratio}}\) in Eq.~\eqref{eq:scaled_ratio}. Fig.~\ref{fig:app_logratio_alpha} reports the ablation on Robomimic Square-state. A moderate value, \(\alpha_{\mathrm{ratio}}=2\), gives the best balance between policy improvement and trust-region stability. Smaller values under-scale the policy ratio and slow down learning, while overly large values do not provide further gain and can reduce optimization stability.

\noindent\textbf{Advantage Weighting.} RoboFFT weights the surrogate policy ratio by the advantage estimate \(\hat{A}_t\), following the PPO objective in Eq.~\eqref{eq:robofft_objective}. To test whether advantage weighting is necessary in our framework, we introduce a variant called RoboFFT w/o AW, which replaces \(\hat{A}_t\) with the corresponding return target. Concretely, we use
\[
    \hat{R}_t = \hat{A}_t + V_{\phi}(o_t),
\]
and normalize it in the same minibatch-wise manner before using it as the ratio weight.

As shown in Fig.~\ref{fig:app_ablation_aw}, removing advantage weighting causes severe instability and performance degradation. This is consistent with the role of advantages in policy-gradient methods: subtracting a state-dependent baseline reduces variance and prevents all high-return states from indiscriminately increasing the likelihood of their sampled actions. In sparse-reward manipulation tasks, this variance reduction is particularly important because successful trajectories can be rare and reward signals are delayed.


\noindent\textbf{PPO-Style Clipped Update.} RoboFFT can be viewed as augmenting advantage-weighted score / flow matching with a PPO-style trust region. To clarify the role of this constraint, we consider RoboFFT w/o PPO, which removes the clipped ratio objective and instead applies a simpler advantage-weighted matching update. This variant is conceptually closer to reward-weighted or energy-weighted flow matching baselines~\citep{ren2024diffusion}, where the current policy is updated directly according to weighted matching losses without a clipped likelihood-ratio surrogate.

For fair comparison, we align learning rates and other optimization hyperparameters with the reward-weighted regression baseline whenever applicable. Fig.~\ref{fig:app_ablation_ppo} shows that the w/o PPO variant suffers from less stable early fine-tuning, while the full RoboFFT objective maintains smoother improvement. The performance gap can shrink in later stages, but the early instability indicates that the clipped ratio is useful for preventing overly aggressive updates induced by the surrogate matching-loss difference.

\subsection{RoboFFT-F Ablations on Transport-State}
\label{app:robofft_f_transport_ablations}

\begin{figure*}[t]
  \centering
  \subfigure[Advantage weighting]{%
    \includegraphics[width=0.31\textwidth]
      {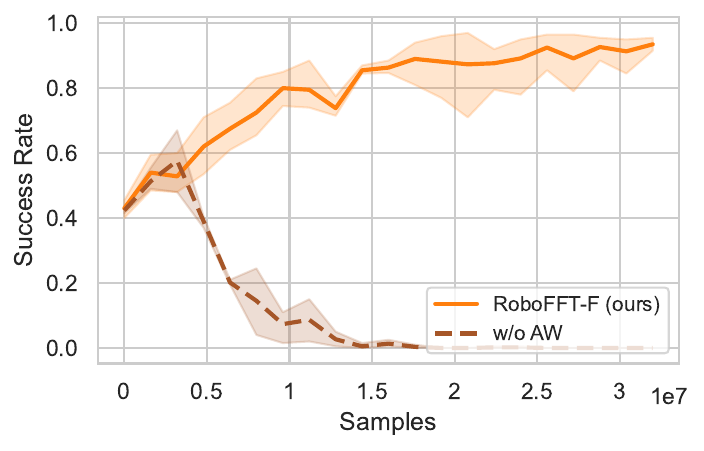}
  }
  \hfill
  \subfigure[Ratio temperature]{%
    \includegraphics[width=0.31\textwidth]
      {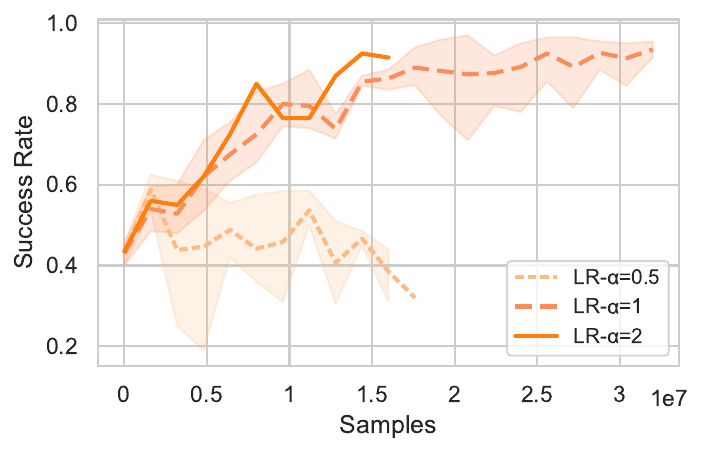}
  }
  \hfill
  \subfigure[MC repetition]{%
    \includegraphics[width=0.31\textwidth]
      {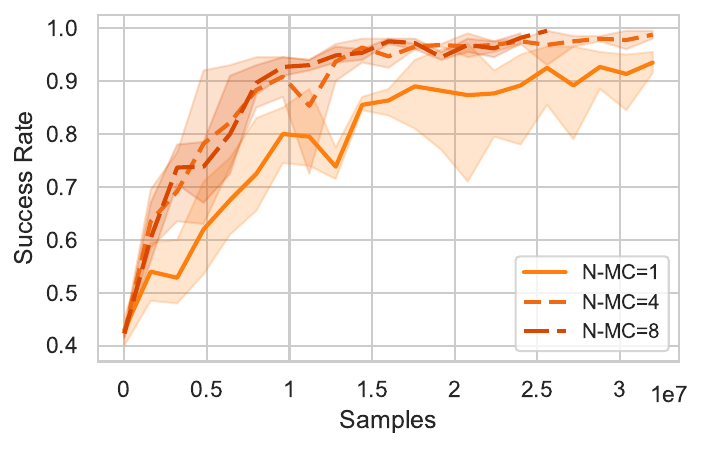}
  }

  \medskip
  \subfigure[PPO-style update]{%
    \includegraphics[width=0.31\textwidth]
      {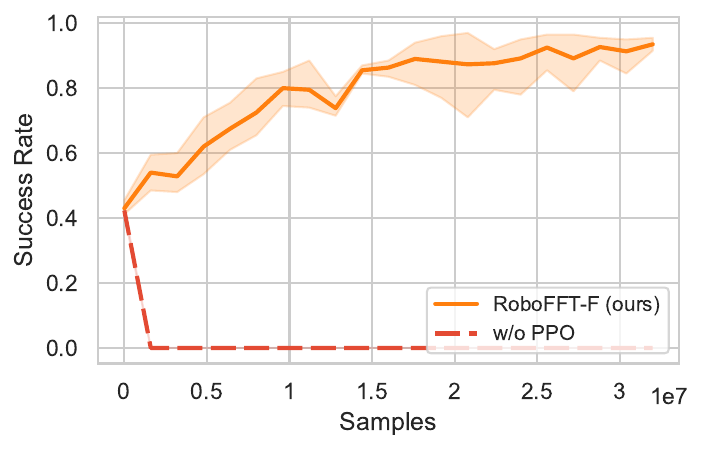}
  }
  \hspace{0.04\textwidth}
  \subfigure[SDE noise strength]{%
    \includegraphics[width=0.31\textwidth]
      {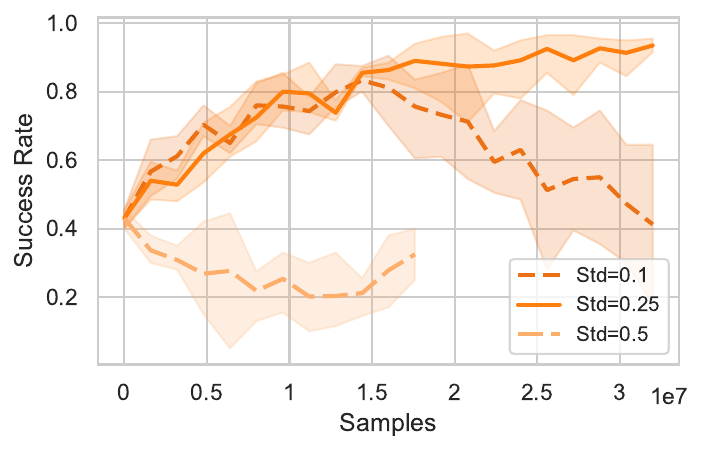}
  }
  \caption{
  RoboFFT-F ablations on Robomimic Transport-state.
  Advantage weighting and the PPO-style update prevent the severe
  degradation observed in the corresponding ablated variants.
  Ratio temperature, MC repetition, and SDE noise also affect
  convergence and stability.
  }
  \label{fig:app_robofft_f_transport_ablations}
\end{figure*}

We repeat the five RoboFFT-F ablations on Transport-state:
advantage weighting, ratio temperature, MC repetition,
the PPO-style update, and SDE noise strength
(Fig.~\ref{fig:app_robofft_f_transport_ablations}).

Replacing advantage weights with return-based weights causes
performance to collapse, consistent with the Square-state result.
The w/o PPO variant also collapses on Transport, revealing a stronger
dependence on the PPO-style surrogate update than on Square,
where the performance gap can narrow later in training.

Ratio temperature again affects stability:
the smallest tested value performs poorly, whereas the larger
tested values support sustained improvement.
Increasing MC repetition from one to four accelerates convergence.
The SDE-noise ablation also favors an intermediate tested noise level,
with both smaller and larger values producing weaker late-stage
performance.

The results from this section and the previous section together support the qualitative relevance and common roles of the
RoboFFT-F design choices across two tasks. Note that this does not imply identical optimal hyperparameters or identical
ablation effects across tasks.

\subsection{RoboFFT-D Ablations on Square-State}
\label{app:robofft_d_ablations}

\begin{figure*}[t]
  \centering
  \subfigure[Negative-advantage discount]{%
    \includegraphics[width=0.31\textwidth]
      {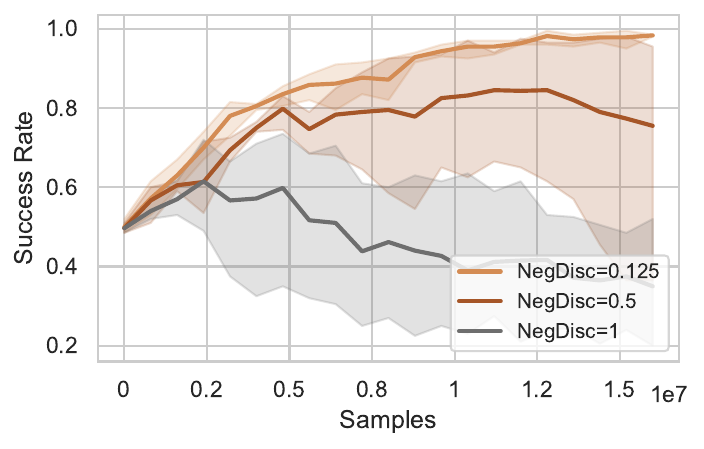}
  }
  \hfill
  \subfigure[Log-SNR weighting]{%
    \includegraphics[width=0.31\textwidth]
      {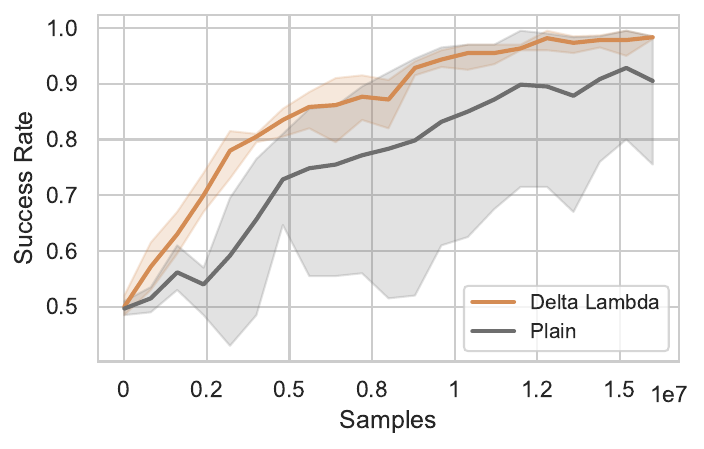}
  }
  \hfill
  \subfigure[Anchor coefficient]{%
    \includegraphics[width=0.31\textwidth]
      {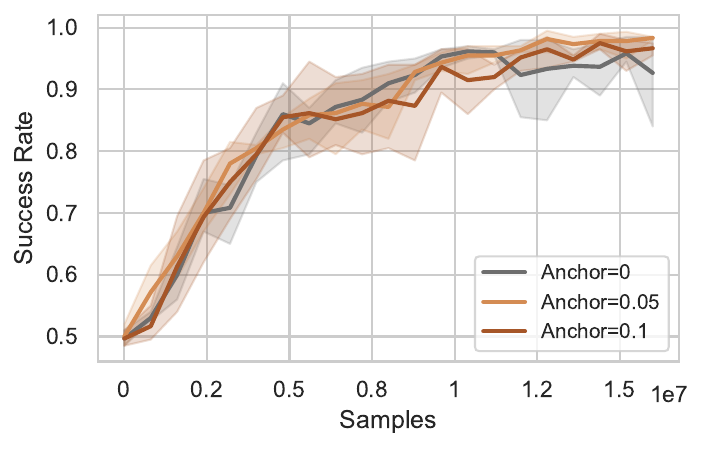}
  }
  \medskip
  \subfigure[MC repetition]{%
    \includegraphics[width=0.31\textwidth]
      {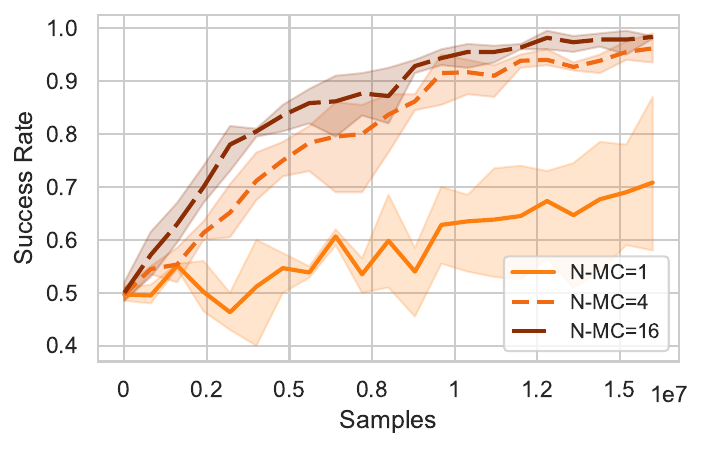}
  }
  \hspace{0.04\textwidth}
  \subfigure[Ratio temperature]{%
    \includegraphics[width=0.31\textwidth]
      {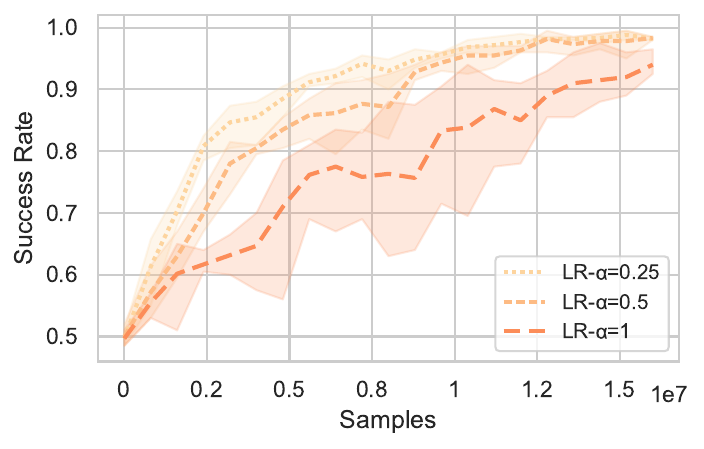}
  }

  \caption{
  RoboFFT-D ablations on Robomimic Square-state.
  Negative-advantage discount, log-SNR weighting and MC repetition substantially affect finetuning performance. Anchor regularization and ratio temperature have comparatively modest effects in the tested configuration.
  }
  \label{fig:app_robofft_d_ablations}
\end{figure*}

Besides the ablations for RoboFFT-F, we examine five design choices in the diffusion-policy implementation RoboFFT-D:
negative-advantage discount, log-SNR weighting, anchor regularization, MC repetition and ratio temperature
(Fig.~\ref{fig:app_robofft_d_ablations}), where the first three are design choices unique to RoboFFT-D.
Their definitions are given in
Appendix~\ref{app:robofft_d_designs}.

\noindent\textbf{Negative-advantage discount.}
Without discount ($\beta_{\mathrm{neg}}=1$), performance deteriorates sharply and finetuning becomes unstable. The tested discounted variants improve stability, with
$\beta_{\mathrm{neg}}=0.125$ providing the strongest sustained
performance. The result suggest a critical impact of negative-advantage discount on the stability of RoboFFT-D.

\noindent\textbf{Log-SNR weighting.}
The $\widetilde{\Delta\lambda}$-weighted surrogate improves convergence
and final performance relative to the plain-weighting variant. This ablation concerns the weighting of denoising errors across
noise levels, rather than the SDE exploration-noise strength
used in RoboFFT-F.

\noindent\textbf{MC repetition.}
Using a single forward-noising sample leads to substantially slower
and weaker improvement than using 4 or 16 samples.
This contrasts with RoboFFT-F on Square-state, where a much smaller
MC budget is sufficient, and supports the larger MC budget used
in the diffusion implementation. We suppose that this is due to the fact that the diffusion policy adopts more stochastic and complicated parametrization compared to the flow policy, thus requiring more repetitions for better likelihood estimation.

\noindent\textbf{Anchor regularization and ratio temperature.}
Compared to without the anchor loss, the anchor coefficients which are $>0$ lead to broadly similar mean success rate on Square-state while slightly reducing the variance, suggesting a relatively modest effect in
this configuration. Ratio temperature affects learning speed, with smaller tested values
producing faster early improvement. These results suggest that these factors have relatively minor impacts on the performance of RoboFFT-D.


\section{Details of Visualization Experiments}
\label{app:visualization_details}

In Sec.~\ref{sec:abl_viz}, we provide a rollout-level visualization on Robomimic Square to understand how RoboFFT changes the action distribution of a pretrained rectified-flow policy. This section describes the full protocol and additionally includes the t-SNE action-embedding visualization that is omitted from the main text for space. We use the same notation as in the main text whenever possible: \(s_t\) denotes the environment state or observation at decision step \(t\), \(a_t\) denotes an action chunk, \(V_{\phi}(s_t)\) is the learned PPO value critic, and \(\pi_{\theta}\) denotes the rectified-flow policy. We write \(\pi_{\mathrm{pre}}\) for the pretrained policy and \(\pi_{\mathrm{ft}}\) for the finetuned policy.

\subsection{Auxiliary Visualization Critic}
\label{app:qviz}

To score sampled action chunks during visualization, we train an auxiliary action-conditioned critic \(Q_{\mathrm{viz}}(s,a)\). This network is used only for analysis and is not used in the actor update. Unlike an optimal-Q critic, \(Q_{\mathrm{viz}}\) is not trained with a Bellman-optimal target or a maximization over next actions. Instead, it is fitted to the one-step policy-Q target induced by the current rollout policy:
\begin{equation}
    y_t^{\mathrm{viz}}
    =
    r_t^{\mathrm{viz}}
    +
    \gamma (1-d_t) V_{\phi}(s_{t+1}),
    \label{eq:app_qviz_target}
\end{equation}
where \(d_t\) is the true termination indicator and \(r_t^{\mathrm{viz}}\) is the reward used for fitting \(Q_{\mathrm{viz}}\). Since \(V_{\phi}\) is trained with reward normalization in our experiments, we use the same normalized reward scale for \(r_t^{\mathrm{viz}}\), so that \(Q_{\mathrm{viz}}\) and \(V_{\phi}\) are numerically aligned. The auxiliary critic is optimized by
\begin{equation}
    \mathcal{L}_{Q_{\mathrm{viz}}}
    =
    \mathbb{E}_{(s_t,a_t,r_t,s_{t+1})}
    \left[
    \left(
        Q_{\mathrm{viz}}(s_t,a_t)-y_t^{\mathrm{viz}}
    \right)^2
    \right].
    \label{eq:app_qviz_loss}
\end{equation}
This makes \(Q_{\mathrm{viz}}\) a diagnostic state-action quality probe for the finetuned policy, rather than a critic used for policy optimization.

For Robomimic Square, the low-dimensional state has dimension \(d_s=23\). We use one observation step and predict action chunks with horizon \(H=4\) and action dimension \(d_a=7\). Thus, \(Q_{\mathrm{viz}}\) takes as input \(s\in\mathbb{R}^{1\times 23}\) and \(a\in\mathbb{R}^{4\times 7}\). The network is implemented as an MLP over the flattened state and flattened action chunk. During fine-tuning, \(Q_{\mathrm{viz}}\) is updated for 4 epochs per iteration, with batch size 4096, learning rate \(10^{-3}\), and gradient clipping with maximum norm 25.

After training, we define the visualization advantage of an action chunk as
\begin{equation}
    A_{\mathrm{viz}}(s,a)
    =
    Q_{\mathrm{viz}}(s,a)-V_{\phi}(s).
    \label{eq:app_aviz}
\end{equation}
This quantity measures the relative quality of an action chunk under a fixed state, using the PPO value critic as a state-dependent baseline.

\subsection{Anchor State Collection}
\label{app:anchor_states}

The t-SNE visualization uses a set of anchor states collected during online fine-tuning. At each fine-tuning iteration, we sample states from the rollout buffer and label them according to whether the corresponding episode succeeds. At each iteration, we collect up to 64 states from successful episodes and 64 states from failed episodes. The collected states are merged across fine-tuning iterations. In the evaluation stage, we sample at most 512 anchor states from this merged set, approximately balancing successful and failed anchors when both are available.

\subsection{Paired Action Sampling}
\label{app:paired_action_sampling}

Both visualizations compare \(\pi_{\mathrm{pre}}\) and \(\pi_{\mathrm{ft}}\) under identical state and latent conditions. Given an anchor state \(s_i\), we sample \(K=64\) initial latent noises
\begin{equation}
    z_{i,k}\sim \mathcal{N}(0,I),
    \qquad
    k=1,\ldots,K.
    \label{eq:app_latent_sampling}
\end{equation}
Let \(\Phi_{\theta}(s,z)\) denote the deterministic rectified-flow sampling map from initial latent \(z\) to the final action chunk under policy parameters \(\theta\). We generate paired action chunks
\begin{equation}
    a^{\mathrm{pre}}_{i,k}
    =
    \Phi_{\theta_{\mathrm{pre}}}(s_i,z_{i,k}),
    \qquad
    a^{\mathrm{ft}}_{i,k}
    =
    \Phi_{\theta_{\mathrm{ft}}}(s_i,z_{i,k}).
    \label{eq:app_paired_sampling}
\end{equation}
Using the same \(z_{i,k}\) for both policies reduces Monte Carlo noise and makes the comparison depend mainly on the learned policy mapping. For each sampled action chunk, we compute \(Q_{\mathrm{viz}}(s_i,a)\), \(A_{\mathrm{viz}}(s_i,a)\), and the sample-based action diversity defined below.

\subsection{Sample-Based Action Diversity}
\label{app:diversity_metric}

Since rectified-flow policies do not provide a convenient closed-form entropy, we use a sample-based diversity metric. For a policy \(\pi\), let
\[
    \mathcal{A}^{\pi}(s)
    =
    \{a_1^\pi(s),\ldots,a_K^\pi(s)\}
\]
be the \(K\) action chunks sampled at state \(s\). Each \(a_k^\pi(s)\in\mathbb{R}^{H\times d_a}\). We define the pair-wise diversity as
\begin{equation}
    D_{\pi}(s)
    =
    \frac{2}{K(K-1)}
    \sum_{1\leq k<\ell\leq K}
    \frac{
        \left\|
        \mathrm{vec}(a_k^\pi(s))
        -
        \mathrm{vec}(a_\ell^\pi(s))
        \right\|_2^2
    }{
        H d_a
    }.
    \label{eq:app_pairwise_diversity}
\end{equation}
The normalization by \(H d_a\) makes the metric less sensitive to the action dimension and action horizon. We use this metric for all diversity curves in Fig.~\ref{fig:viz_rollout}.

\subsection{Rollout-Timeline Visualization}
\label{app:rollout_timeline}

For Fig.~\ref{fig:viz_rollout}, we first collect rollouts using the finetuned policy \(\pi_{\mathrm{ft}}\) in Robomimic Square. We run 20 parallel environments for at most 400 decision steps. Among completed episodes, we retain successful ones. To select a representative trajectory that contains both pre-success and post-success phases, we compute the positive-reward ratio
\begin{equation}
    \rho
    =
    \frac{1}{T}
    \sum_{t=1}^{T}
    \mathbf{1}[r_t>0],
    \label{eq:app_positive_reward_ratio}
\end{equation}
where \(T\) is the trajectory length in decision steps. We choose the successful trajectory minimizing
\begin{equation}
    S_{\mathrm{traj}}
    =
    |\rho-0.5|
    -
    0.01 \cdot \mathrm{Std}(\{r_t\}_{t=1}^{T}).
    \label{eq:app_trajectory_selection}
\end{equation}
This favors trajectories whose reward transition occurs near the middle of the rollout while avoiding completely flat reward traces.

For every state \(s_t\) along the selected trajectory, we repeat the paired action sampling in Eq.~\eqref{eq:app_paired_sampling}. We then compute the action-value gap between the finetuned and pretrained policies:
\begin{equation}
    \Delta Q_{\mathrm{viz}}(s_t)
    =
    \frac{1}{K}
    \sum_{k=1}^{K}
    Q_{\mathrm{viz}}(s_t,a^{\mathrm{ft}}_{t,k})
    -
    \frac{1}{K}
    \sum_{k=1}^{K}
    Q_{\mathrm{viz}}(s_t,a^{\mathrm{pre}}_{t,k}).
    \label{eq:app_delta_qviz}
\end{equation}
We also compute
\begin{equation}
    D_{\mathrm{pre}}(s_t)=D_{\pi_{\mathrm{pre}}}(s_t),
    \qquad
    D_{\mathrm{ft}}(s_t)=D_{\pi_{\mathrm{ft}}}(s_t),
    \label{eq:app_timeline_diversity}
\end{equation}
and their difference
\begin{equation}
    \Delta D(s_t)
    =
    D_{\mathrm{pre}}(s_t)-D_{\mathrm{ft}}(s_t).
    \label{eq:app_diversity_reduction}
\end{equation}
Positive \(\Delta D(s_t)\) indicates that the finetuned policy produces a more concentrated sampled action distribution at \(s_t\). The rollout-timeline visualization plots \(r_t\), \(\Delta Q_{\mathrm{viz}}(s_t)\), \(D_{\mathrm{pre}}(s_t)\), \(D_{\mathrm{ft}}(s_t)\), and \(\Delta D(s_t)\) over decision steps. The main-text observation that \(\Delta Q_{\mathrm{viz}}\) increases while diversity decreases around key decision stages indicates that RoboFFT guides the policy toward more concentrated and higher-quality action samples.

\subsection{t-SNE Action-Embedding Visualization}
\label{app:tsne_visualization}

For the t-SNE visualization, we use the collected anchor states rather than a single trajectory. For each selected anchor state, we sample \(K=64\) action chunks from both \(\pi_{\mathrm{pre}}\) and \(\pi_{\mathrm{ft}}\) using the same latent noises. Each action chunk is flattened into a vector in \(\mathbb{R}^{H d_a}\), which is 28-dimensional for Robomimic Square.

To make the embedding balanced and computationally manageable, we randomly subsample at most 2500 action chunks from each policy. We then form a joint action set
\begin{equation}
    \mathcal{X}
    =
    \left\{
    \mathrm{vec}(a^{\mathrm{pre}}_{i,k})
    \right\}_{i,k}
    \cup
    \left\{
    \mathrm{vec}(a^{\mathrm{ft }}_{i,k})
    \right\}_{i,k}.
    \label{eq:app_joint_action_set}
\end{equation}
A single two-dimensional embedding is computed on the joint set \(\mathcal{X}\), and the resulting points are split into pretrained and finetuned panels for visualization. Before t-SNE, we mean-center \(\mathcal{X}\) and apply PCA preprocessing to at most 30 dimensions. We then run t-SNE with perplexity 35 and 1000 optimization iterations. If t-SNE is unavailable or fails, we fall back to a two-dimensional PCA projection.

Each point in Fig.~\ref{fig:app_viz_act_embed} corresponds to one sampled action chunk and is colored by \(A_{\mathrm{viz}}(s,a)\) from Eq.~\eqref{eq:app_aviz}. The pretrained and finetuned panels use separate blue and orange color maps but share the same color range. To reduce the influence of outliers, the color limits are determined by the 98-th percentile range of all \(A_{\mathrm{viz}}\) values from both policies. In the default setting, we do not subtract anchor-wise action means before t-SNE; therefore, the embedding visualizes the global action manifold across both state-dependent action variations and task-phase differences. 

\subsection{Interpretation of the t-SNE Visualization}
\label{app:tsne_interpretation}

\begin{figure*}[!th]
  \centering
  \includegraphics[width=0.86\textwidth]{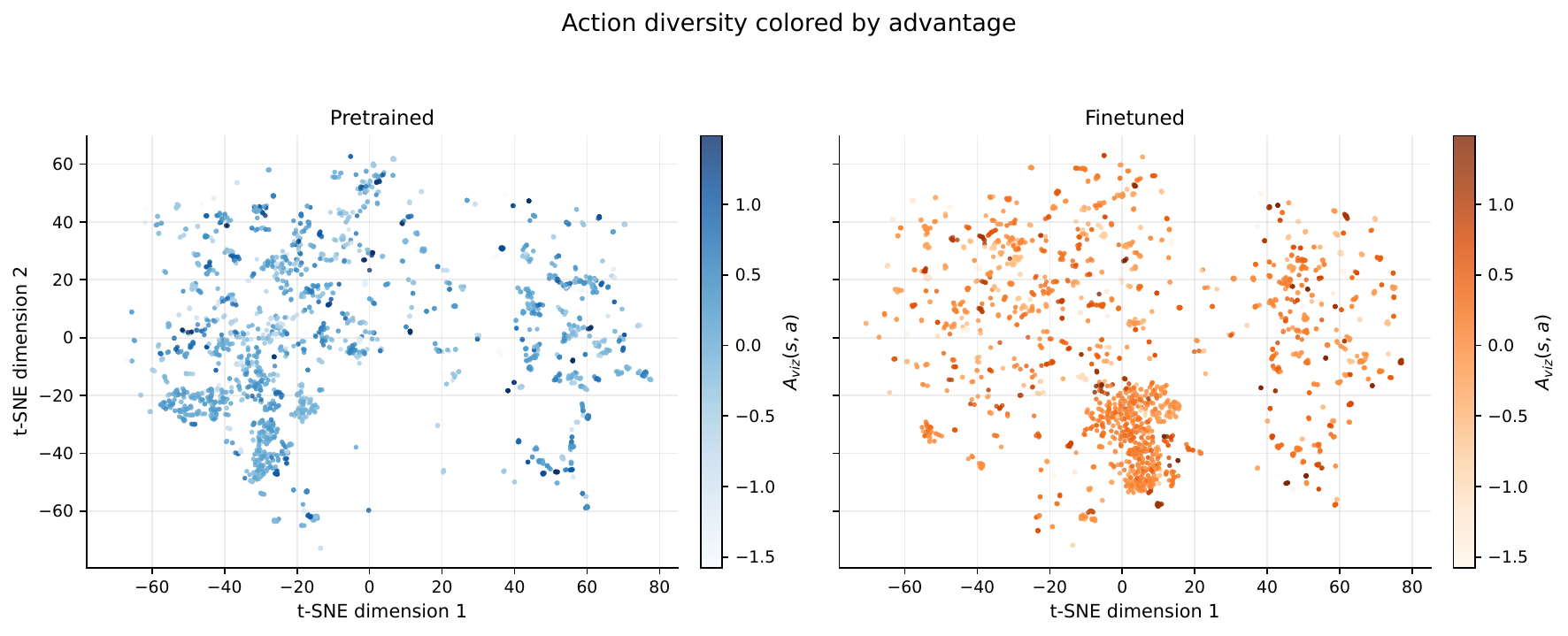}
  \caption{
  t-SNE visualization of action chunks sampled from the pretrained and finetuned policies under matched anchor states and latent noises. Each point is colored by the diagnostic advantage \(A_{\mathrm{viz}}(s,a)=Q_{\mathrm{viz}}(s,a)-V_{\phi}(s)\). The finetuned policy forms a denser region of higher-advantage actions, complementing the rollout-timeline analysis.
  }
  \label{fig:app_viz_act_embed}
\end{figure*}

The t-SNE visualization in Fig.~\ref{fig:app_viz_act_embed} provides a complementary view of how RoboFFT reshapes the action distribution beyond a single rollout trajectory. Compared with the pretrained policy, whose sampled action chunks are more diffusely distributed and contain many low- or negative-advantage samples, the finetuned policy forms a more concentrated region with noticeably higher diagnostic advantages. In particular, after fine-tuning, a dense cluster of positive-advantage actions emerges in the embedding space, indicating that RoboFFT increases the probability mass around actions that are preferred by the diagnostic critic \(Q_{\mathrm{viz}}\).

This observation is consistent with the rollout-timeline analysis in Fig.~\ref{fig:viz_rollout}. The timeline visualization shows that, at key decision stages such as grasping, transporting, and aligning the object, the finetuned policy improves the average action quality while reducing sample diversity. The t-SNE visualization further shows that this effect is not restricted to one selected trajectory: across a broader set of anchor states collected from successful and failed rollouts, the finetuned policy tends to move sampled action chunks toward higher-advantage regions. Together, these results suggest that RoboFFT improves the pretrained generative policy by concentrating its action distribution around more task-relevant and higher-quality action modes, rather than merely increasing random exploration. Note that the density of a t-SNE embedding is an intuitive signal of action mode concentration instead of a direct estimate of
policy likelihood or the number of behavioral modes. The observed concentration therefore does not by itself establish
either mode collapse or preservation of all useful modes.

\section{Implementation details for VLA }
\label{app:vla_details}

We use RLinf~\cite{yu2026rlinf} for implementing \textbf{RoboFFT} for VLA finetuning. 
Specifically, we only replace the calculation of log probability with the surrogate matching loss. The data collection and PPO update stay the same. 
Tab.\ref{tab:robofft_pirl_config_diff} demonstrates the configuration difference between RoboFFT and $\pi_{RL}$~\cite{chen2025pirl}.

\begin{table}[!th]
\centering
\small
\caption{Differences in training configurations  between RoboFFT and  $\pi_{RL}$  baseline.}
\begin{tabular}{lcc}
\toprule
\textbf{Parameter} & \textbf{RoboFFT} & \textbf{$\pi_{RL}$} \\
\midrule


Log probability calculation
& score/ flow matching loss
& multi step MDP \\

KL coefficient $\beta$ 
& 0.5 
& 0.0 \\

PPO clipping range 
& 0.05 / 0.05 
& 0.20 / 0.20 \\

Samples per action $N_\mathrm{mc}$
& 8 
& Not used \\


\bottomrule
\end{tabular}
\label{tab:robofft_pirl_config_diff}
\end{table}

\section{Real World Experiments}
\label{app:real_details}
\subsection{Real world settings}

\begin{figure}[!h]
  \centering

      \includegraphics[width=1.0\textwidth]{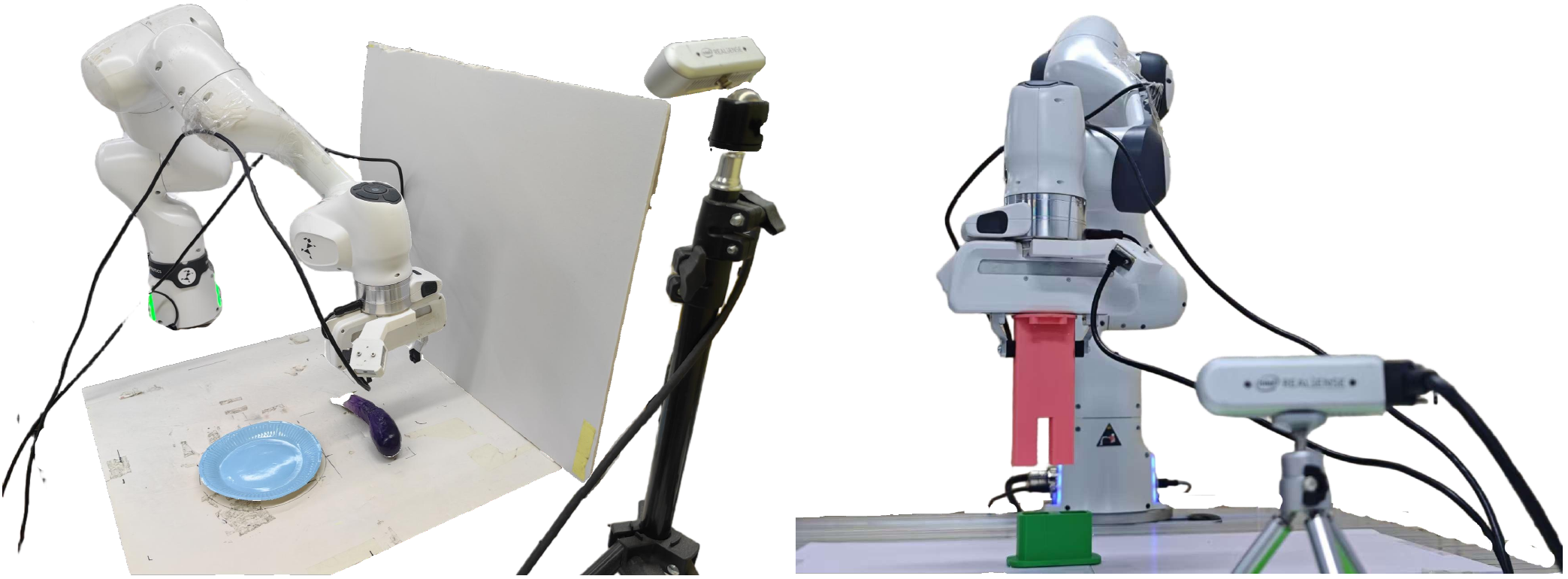}
  \caption{
  Real world settings 
  }
  \label{fig:settings}
  
\end{figure}

Fig.\ref{fig:settings} demonstrates the  settings of our two real world platforms. 
For \textit{Pick Eggplant} and \textit{Cabinet Retrieval} task, we use Franka Research 3. Two wrist Realsense D405 cameras are mounted on the end effector to capture fine-grained manipulation details, where a third-view camera is employed for whole scene observation. 
For \textit{Peg Insertion} and \textit{Pour}, we use Franka Emika Panda. The camera settings are the same except for the position of the third-view camera. 
Tab.~\ref{tab:real_task_suite} and Tab.~\ref{tab:real_state_action} summarize the task configurations and observation/action spaces, respectively..

\begin{table}[t]
\centering
\caption{Real-world task suite and key challenges.}
\label{tab:real_task_suite}

\small
\setlength{\tabcolsep}{2.5pt}
\renewcommand{\arraystretch}{1.15}

\begin{tabularx}{\columnwidth}{
    >{\hsize=0.82\hsize\raggedright\arraybackslash}X
    >{\hsize=0.90\hsize\raggedright\arraybackslash}X
    >{\hsize=0.82\hsize\raggedright\arraybackslash}X
    >{\hsize=1.15\hsize\raggedright\arraybackslash}X
    >{\hsize=1.31\hsize\raggedright\arraybackslash}X
}
\toprule
\textbf{Task} &
\textbf{Embodiment} &
\textbf{Task Type} &
\textbf{Randomization} &
\textbf{Key Challenges} \\
\midrule

Pick Eggplant &
Franka Research 3 &
Pick-and-place &
Object position randomized within $3\,\mathrm{cm}$ along $x/y$ &
Grasping an irregular object under position variation; stable transport and accurate placement. \\

Peg Insertion &
Franka Emika Panda &
Precision assembly &
Initial end-effector position randomized within $1.5\,\mathrm{cm}$ along $x/y$ &
Fine-grained pose alignment, tight insertion tolerance, and contact-rich manipulation. \\

Pouring &
Franka Emika Panda &
Granular &
Object position randomized within $1\,\mathrm{cm}$ along $x/y$  &
Continuous orientation control, target alignment, and error propagation across sequential stages. \\

Cabinet Retrieval &
Franka Research 3 &
Long-horizon &
Object position randomized within $1\,\mathrm{cm}$ along $x/y$ &
Long-horizon control, confined-space manipulation, partial occlusion, and sensitivity to intermediate failures. \\

\bottomrule
\end{tabularx}
\end{table}







\begin{table}[t]
\centering
\caption{State and action components of the real-world tasks.
We abbreviate dimension as ``dim.'' and proprioception as ``prop.''.}
\label{tab:real_state_action}

\small
\setlength{\tabcolsep}{2.5pt}
\renewcommand{\arraystretch}{1.15}

\begin{tabularx}{\columnwidth}{
    >{\hsize=0.90\hsize\raggedright\arraybackslash}X
    >{\hsize=1.15\hsize\centering\arraybackslash}X
    >{\hsize=0.55\hsize\centering\arraybackslash}X
    >{\hsize=1.25\hsize\raggedright\arraybackslash}X
    >{\hsize=0.45\hsize\centering\arraybackslash}X
    >{\hsize=1.70\hsize\raggedright\arraybackslash}X
}
\toprule
\textbf{Task} &
\multicolumn{3}{c}{\textbf{Observation Space}} &
\multicolumn{2}{c}{\textbf{Action Space}} \\
\cmidrule(lr){2-4}
\cmidrule(lr){5-6}

&
\textbf{Visual Obs.} &
\textbf{Prop. dim.} &
\textbf{Prop. State} &
\textbf{Dim.} &
\textbf{Action Representation} \\
\midrule

Pick Eggplant &
RGB $(224,224,3)\times3$ &
8 &
7-D joint state + 1-D gripper state &
7 &
6-D target end-effector pose + 1-D gripper command \\

Peg Insertion &
RGB $(224,224,3)\times3$ &
7 &
7-D joint state &
6 &
6-D target end-effector pose; gripper fixed during insertion \\

Pouring &
RGB $(224,224,3)\times3$ &
8 &
7-D joint state + 1-D gripper state &
7 &
6-D target end-effector pose + 1-D gripper command \\

Cabinet Retrieval &
RGB $(224,224,3)\times3$ &
8 &
7-D joint state + 1-D gripper state &
7 &
6-D target end-effector pose + 1-D gripper command \\

\bottomrule
\end{tabularx}
\end{table}








\subsection{Training Details}

We adapt \textbf{RoboFFT} into a practical offline-to-online training paradigm for deployment on physical robots (Algo.~\ref{alg:real_robofft}).
For each real-world task, we first collect $N_{\mathrm{demo}}=50$ human
demonstrations $\mathcal{D}_{\mathrm{demo}}$ using SpaceMouse
teleoperation. A rectified-flow policy $u_{\theta}$ is pretrained on
$\mathcal{D}_{\mathrm{demo}}$ via imitation learning, while the critic
is pretrained using IQL~\citep{kostrikov2021offline}. We adopt
rectified flow as the base generative policy because its efficient
sampling enables practical real-time execution on physical robots.
We use a sparse binary reward
$r\in\{0,1\}$ based on task completion, which is directly judged by a
human evaluator without training an additional reward model. 

\noindent\textbf{Offline RL.} 
Starting from the pretrained actor and critic, we first perform offline
\textbf{RoboFFT} optimization using the available offline trajectories.
Since the training data are not collected by the current policy, we use
the IQL critic to estimate the offline advantage
$\hat{A}_t = Q_{\psi}(o_t,a_t^0)-V_{\phi}(o_t)$.
The estimated advantages are then used in the same forward process
policy optimization objective as in the main \textbf{RoboFFT}
algorithm. Specifically, we perturb the sampled actions through forward
noising, evaluate the matching losses under the current and reference
policies, construct the surrogate policy ratio, and optimize the
PPO-style clipped objective. Thus, the algorithmic core of
\textbf{RoboFFT} remains unchanged, while IQL provides the advantage
estimates required for offline policy optimization.

\noindent\textbf{Real-world rollout and policy aggregation.}
After offline RL, we deploy the resulting policy on the physical robot and collect \(N_{\mathrm{roll}}\) rounds of rollout trajectories, denoted as \(\mathcal{D}_{\mathrm{roll}}\), where each round contains \(d\) trajectories.
To stabilize subsequent online optimization, we combine the newly
collected trajectories with the original human demonstrations,
$\mathcal{D}_{\mathrm{agg}}
=\mathcal{D}_{\mathrm{demo}}\cup\mathcal{D}_{\mathrm{roll}}$,
and perform imitation learning on the aggregated dataset.
This step distills the improved behaviors discovered by offline RL and
real-world interaction while preserving reliable behaviors from the
human demonstrations, thereby reducing policy drift before online finetuning. We normally set $N_{\mathrm{roll}}=1$ and $d=30$.

\noindent\textbf{Online RL.}
The policy obtained from data aggregation is used to initialize the
online actor, while the pretrained IQL critic provides the initialization
for the online value function. We then perform online \textbf{RoboFFT}
for $K_{\mathrm{on}}$ iterations. At each iteration, the current policy
is deployed to collect a batch $B_{\mathrm{on}}$ of  on-policy
trajectories, from which advantage estimates $\hat{A}_t$ and value
targets $\hat{R}_t$ are computed. The actor is optimized using the same
forward-noising surrogate ratio and PPO-style clipped objective as in
the simulation experiments, while the value function is updated using
the newly collected on-policy data. After each update, the new policy is
deployed for the next interaction round. We normally set $K_{\mathrm{on}} = 5$.

\begin{algorithm}[tb]
\caption{Real-World Offline-to-Online RoboFFT}
\label{alg:real_robofft}
\begin{algorithmic}[1]
\Require Pretrained prediction network $u_{\theta}$,
human demonstrations $\mathcal{D}_{\mathrm{demo}}$,
offline dataset $\mathcal{D}_{\mathrm{off}}$,
IQL critic $Q_{\psi}$ and value function $V_{\phi}$,
number of Monte Carlo samples $N_{\mathrm{mc}}$,
ratio temperature $\alpha_{\mathrm{ratio}}$,
PPO clipping threshold $\epsilon_{\mathrm{clip}}$.

\State Pretrain $u_{\theta}$ on $\mathcal{D}_{\mathrm{demo}}$ via imitation learning.
\State Deploy $u_{\theta}$ to collect initial real-world rollouts
and initialize $\mathcal{D}_{\mathrm{off}}$.

\For{each offline RL iteration}
    \State Train $Q_{\psi}$ and $V_{\phi}$ with IQL on
    $\mathcal{D}_{\mathrm{off}}$.
    \State Compute offline advantages
    $\hat{A}_t=Q_{\psi}(o_t,a_t^0)-V_{\phi}(o_t)$.

    \State Set $\theta_{\mathrm{old}}\leftarrow\theta$.
    \For{each policy update epoch}
        \For{each minibatch from $\mathcal{D}_{\mathrm{off}}$}
            \State Draw $\{(k_i,\epsilon_i)\}_{i=1}^{N_{\mathrm{mc}}}$,
            where $k_i\sim\mathcal{U}(0,1)$ and
            $\epsilon_i\sim\mathcal{N}(0,I)$.
            \State Construct noised actions
            $a_t^{k_i}=\alpha_{k_i}a_t^0+\sigma_{k_i}\epsilon_i$.
            \State Compute
            $\widehat{\mathcal{L}}_{\theta}^{w}(a_t^0,o_t)$ and
            $\widehat{\mathcal{L}}_{\theta_{\mathrm{old}}}^{w}(a_t^0,o_t)$
            using Eq.~\eqref{eq:mc_matching_loss}.
            \State Compute
            $\widehat{\rho}^{\,s}_t(\theta)$
            using Eq.~\eqref{eq:scaled_ratio}.
            \State Update $\theta$ by maximizing
            Eq.~\eqref{eq:robofft_objective}
            with offline advantage $\hat{A}_t$.
        \EndFor
    \EndFor

    \State Deploy the updated policy and collect new real-world rollouts
    $\mathcal{D}_{\mathrm{roll}}$.
    \State Update
    $\mathcal{D}_{\mathrm{off}}
    \leftarrow
    \mathcal{D}_{\mathrm{off}}\cup\mathcal{D}_{\mathrm{roll}}$.
    \State Perform imitation learning on
    $\mathcal{D}_{\mathrm{demo}}\cup\mathcal{D}_{\mathrm{roll}}$
    to obtain the base policy for the next iteration.
\EndFor

\State Initialize the online actor with the offline-trained $u_{\theta}$
and the critic with the IQL value function $V_{\phi}$.

\For{each online RL iteration}
    \State Set $\theta_{\mathrm{old}}\leftarrow\theta$.
    \State Collect on-policy real-world rollouts with $u_{\theta}$ and
    store $(o_t,a_t^0,r_t,o_{t+1})$ in $\mathcal{B}$.
    \State Compute advantage estimates $\hat{A}_t$ and value targets
    $\hat{R}_t$.

    \For{each policy update epoch}
        \For{each minibatch from $\mathcal{B}$}
            \State Draw $\{(k_i,\epsilon_i)\}_{i=1}^{N_{\mathrm{mc}}}$ with
            $k_i\sim\mathcal{U}(0,1)$ and
            $\epsilon_i\sim\mathcal{N}(0,I)$.
            \State Construct
            $a_t^{k_i}=\alpha_{k_i}a_t^0+\sigma_{k_i}\epsilon_i$.
            \State Compute
            $\widehat{\mathcal{L}}_{\theta}^{w}$ and
            $\widehat{\mathcal{L}}_{\theta_{\mathrm{old}}}^{w}$
            using Eq.~\eqref{eq:mc_matching_loss}.
            \State Compute
            $\widehat{\rho}^{\,s}_t(\theta)$
            using Eq.~\eqref{eq:scaled_ratio}.
            \State Update $\theta$ by maximizing
            Eq.~\eqref{eq:robofft_objective}.
            \State Update $V_{\phi}$ by minimizing
            Eq.~\eqref{eq:app_value_loss}.
        \EndFor
    \EndFor
    \State Clear $\mathcal{B}$.
\EndFor

\State \Return Fine-tuned generative policy $u_{\theta}$.
\end{algorithmic}
\end{algorithm}

\section{Extended Related Work}

\subsection{Generative Models for Robot Manipulation}

Generative models have become an increasingly important policy class for robot manipulation, as they provide expressive distributions over complex and potentially multimodal action sequences. Early approaches~\citep{zhao2023learning, li2024unidoormanip} formulate imitation learning with a conditional variational autoencoder (CVAE). More recently, diffusion-based policies have demonstrated substantially greater expressiveness for continuous control. Diffusion Policy~\citep{chi2025diffusion, janner2022planning} models the conditional action distribution through an iterative denoising process, enabling stable learning of high-dimensional and multimodal action distributions and motivating a broad family of diffusion-based manipulation policies. Flow-matching models provide an alternative generative formulation by learning continuous transport from a simple noise distribution to the action distribution. Methods such as ManiFlow~\citep{yan2025maniflow} further improve sampling efficiency and policy expressiveness through consistency-based flow training, demonstrating strong performance on diverse manipulation tasks. In parallel, vision-language-action (VLA) models have extended robot policies toward large-scale generalist control by leveraging pretrained vision-language representations and heterogeneous robot datasets. RT-2~\citep{brohan2023rt} and OpenVLA~\citep{kim2024openvla} generate discretized robot actions within large autoregressive vision-language models, while $\pi_0$~\citep{black2024pi} and $\pi_{0.5}$~\citep{intelligence2025pi} combine a pretrained vision-language backbone with a flow-matching action expert for continuous action generation. 

\subsection{RL for Generative Robot Policy}

Reinforcement learning has been increasingly explored to improve generative policies beyond behavior cloning. Early studies mainly focus on offline RL with fixed datasets. Diffusion Q Learning~\citep{wang2022diffusion} parameterizes the actor as a diffusion model and combines diffusion regression with Q value maximization, while IDQL~\citep{hansen2023idql} uses an implicit Q-function to select actions sampled from a diffusion behavior policy. This idea has also been extended to flow-based policies. Flow Q Learning (FQL)~\citep{park2025flow} learns a flow matching behavior policy together with a one-step RL policy, avoiding recursive differentiation through iterative flow generation. Although these methods exploit the expressiveness of generative policies, their improvement is constrained by the coverage of the offline dataset and the accuracy of learned value functions.

More recent work considers online or offline-to-online adaptation of pretrained generative robot policies. One direction keeps the base policy frozen and learns a lightweight steering or residual component. DSRL~\citep{wagenmaker2025steering} performs RL in the latent-noise space of a pretrained diffusion or flow policy without modifying its parameters, while residual RL methods learn corrective actions on top of a frozen behavior cloning policy~\citep{ankile2025residual, sun2026prior, jiang2024transic}. These approaches preserve pretrained behaviors and reduce the complexity of online optimization, but restrict policy improvement to the residual or latent space exposed by the frozen policy.

In contrast, full policy finetuning directly updates the generative model parameters. DPPO~\citep{ren2024diffusion} formulates diffusion denoising as a multi-step MDP and applies PPO over reverse denoising transitions. ReinFlow~\citep{zhang2025reinflow} extends likelihood-based online optimization to flow policies by injecting stochasticity into deterministic flow trajectories. Forward process methods such as FPO~\citep{mcallister2025flow} and its extensions~\citep{yi2026flow} instead construct policy-gradient objectives from flow matching losses, avoiding explicit marginal likelihood computation. Reinforced finetuning has also been extended to VLA policies, with ~\citep{chen2025conrft, xu2026rl} combining offline value learning and online policy optimization for real-world manipulation. RoboFFT belongs to the full policy finetuning family, but performs PPO-style optimization through the forward noising process, providing a unified framework for diffusion, flow, and VLA policies.


\end{document}